\PassOptionsToPackage{shortlabels}{enumitem}
\documentclass[11pt, a4paper]{include/gdm_format}
\usepackage[authoryear, sort&compress, round]{natbib}
\usepackage{xcolor}
\usepackage{subcaption}
\usepackage{mwe}
\usepackage{graphicx}
\usepackage{xr}
\usepackage{marvosym}
\usepackage{makecell}

\usepackage{fvextra}

\definecolor{bgcolor}{rgb}{0.95,0.95,0.95}

\definecolor{lblA}{RGB}{33,113,181}
\definecolor{lblB}{RGB}{230,85,13}

\usepackage{tablefootnote}
\usepackage{pdfpages}
\usepackage{etoc}

\usepackage{hyperref}
\usepackage{url}
\usepackage{xurl}
\usepackage[parfill]{parskip}

\usepackage{amsmath}
\usepackage{xspace}
\usepackage{multirow}
\usepackage{paracol}
\usepackage{booktabs}
\usepackage{color}
\usepackage{colortbl}
\usepackage{cleveref}
\usepackage{graphicx}
\usepackage{algorithm}
\usepackage{algorithmicx}
\usepackage{algpseudocode}
\usepackage{subcaption}
\usepackage{wrapfig}
\usepackage{sidecap}
\usepackage{soul}
\sidecaptionvpos{figure}{t}
\usepackage{multicol}
\usepackage{pifont}
\usepackage[normalem]{ulem}
\usepackage{listings}
\usepackage{changes}

\newcommand{\ouralgo}{\textsc{ShinkaEvolve}\xspace}

\title{\ouralgo: Towards Open-Ended \textit{And} Sample-Efficient Program Evolution}

\correspondingauthor{Robert Tjarko Lange (robert@sakana.ai), Edoardo Cetin (edo@sakana.ai). Robert initiated and led the project; Robert and Edoardo were core code contributors. See \Cref{sec:author_contribution_list} for detailed information on project contributions.}

\author{Robert Tjarko Lange, Yuki Imajuku and Edoardo Cetin\\Sakana AI}

\usepackage{tcolorbox}
\tcbuselibrary{breakable}
\tcbuselibrary{listings}  
\definecolor{lightgreen}{RGB}{235, 255, 235}  

\lstdefinelanguage{CUDA}{
    morekeywords={
        __global__, __device__, __host__, __shared__, __constant__,
        dim3, gridDim, blockDim, blockIdx, threadIdx, syncthreads
    },
    sensitive=true,
    morecomment=[l]{//},
    morecomment=[s]{/*}{*/},
    morestring=[b]"
}

\definecolor{codegreen}{rgb}{0,0.6,0}
\definecolor{codegray}{rgb}{0.5,0.5,0.5}
\definecolor{codepurple}{rgb}{0.58,0,0.82}
\definecolor{backcolour}{RGB}{245,248,250}
\definecolor{emph}{RGB}{166,88,53}
\definecolor{nightblue}{RGB}{9,49,105}
\definecolor{keywords}{RGB}{207,33,46}
\definecolor{lightpurple}{RGB}{130,81,223}

\lstdefinestyle{mystyle}{
    backgroundcolor=\color{backcolour},   
    commentstyle=\color{codegreen},
    keywordstyle=\color{keywords},
    stringstyle=\color{nightblue},
    basicstyle=\ttfamily\scriptsize,
    breakatwhitespace=false,         
    breaklines=true,                 
    captionpos=b,                    
    keepspaces=true,                 
    showspaces=false,                
    showstringspaces=false,
    showtabs=false,                  
    tabsize=2,
    frame=shadowbox,
    emph={AutoTokenizer,AutoModelForSequenceClassification,Explainer},
    emphstyle={\color{emph}},
    emph={[2]from_pretrained,compute_table},
    emphstyle={[2]\color{lightpurple}},
}

\usepackage{CJKutf8}
\begin{document}

\begin{abstract}
We introduce \ouralgo\footnotemark: a new open-source framework leveraging large language models (LLMs) to advance scientific discovery with state-of-the-art performance and unprecedented efficiency. 
Recent advances in scaling inference time compute of LLMs have enabled significant progress in generalized scientific discovery. These approaches rely on evolutionary agentic harnesses that leverage LLMs as mutation operators to generate candidate solutions. 
However, current code evolution methods suffer from critical limitations: they are sample inefficient, requiring thousands of samples to identify effective solutions, and remain closed-source, hindering broad adoption and extension.
\ouralgo addresses these limitations, introducing three key innovations: a parent sampling technique balancing exploration and exploitation, code novelty rejection-sampling for efficient search space exploration, and a bandit-based LLM ensemble selection strategy.
We evaluate \ouralgo across diverse tasks, demonstrating consistent improvements in sample efficiency and solution quality. \ouralgo discovers a new state-of-the-art circle packing solution using only 150 samples, designs high-performing agentic harnesses for AIME mathematical reasoning tasks, identifies improvements to ALE-Bench competitive programming solutions, and discovers novel mixture-of-expert load balancing loss functions that illuminate the space of optimization strategies.
Our results demonstrate that \ouralgo achieves broad applicability with exceptional sample efficiency. By providing open-source accessibility and cost-efficiency, this work democratizes open-ended discovery across diverse computational problems.

\vspace{0.5em}
\begin{center}
\begin{tabular}{rcl}
\raisebox{-1.5pt}{\includegraphics[height=1.05em]{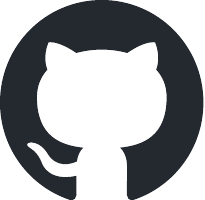}} & \textbf{Code} & \href{https://github.com/SakanaAI/ShinkaEvolve}{\path{https://github.com/SakanaAI/ShinkaEvolve}} \\
\end{tabular}
\end{center}
\end{abstract}

\maketitle
\setcounter{tocdepth}{2}
\etocdepthtag.toc{mtchapter}
\etocsettagdepth{mtchapter}{subsection}
\etocsettagdepth{mtappendix}{none}

\footnotetext{The Japanese term `shinka' translates to `evolution' or `innovation'. \ouralgo refers to the vision of an open-ended self-refining innovation engine. While \ouralgo might be a bit funny for some Japanese readers, as it sounds like Evolve-Evolve or \begin{CJK}{UTF8}{min}進化ー進化\end{CJK}, similar bilingual repetitive terms appear quite often in Japan.}

\begin{figure}[h!]
    \centering
    \includegraphics[width=0.925\textwidth]{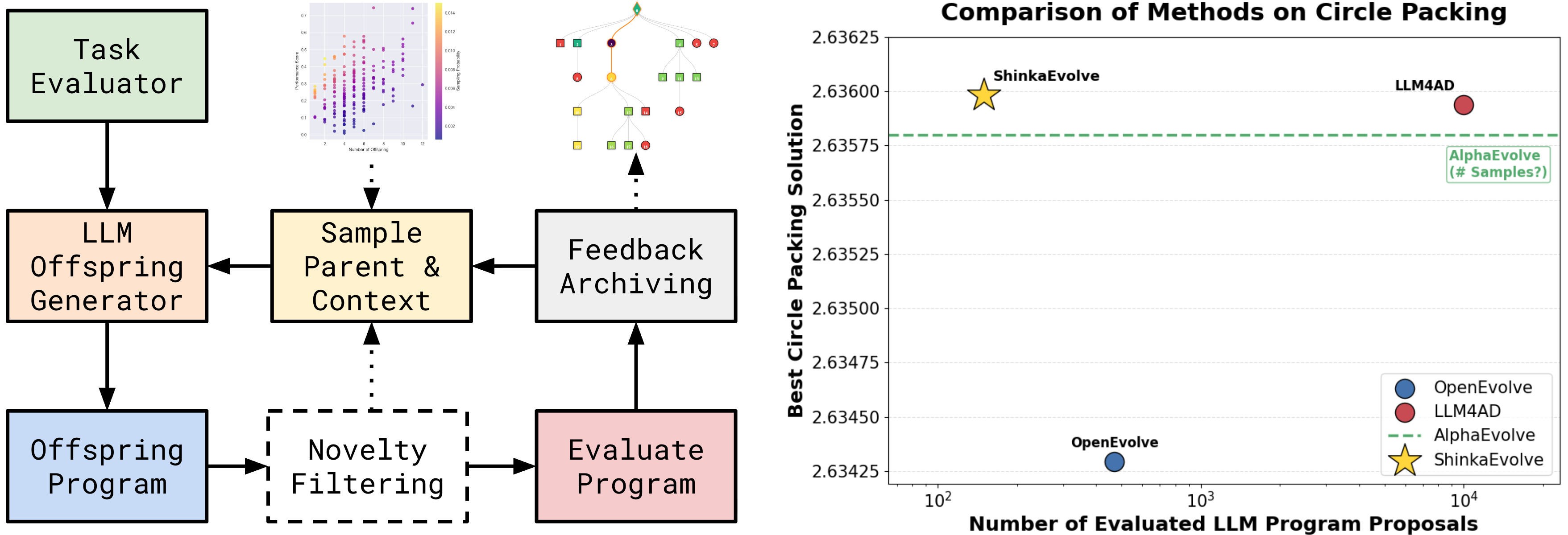}
    \caption{\textbf{High-level overview of \ouralgo.} \textit{Left:} The \ouralgo framework constructs an archive of evaluated programs, rejection-samples new programs, and evaluates their fitness. \textit{Right:} \ouralgo provides a sample efficient alternative to AlphaEvolve and outperforms its Circle Packing solution.}
    \label{fig:conceptual}
\end{figure}

\section{Introduction}

The rapid advancement of large language models (LLMs) has transformed scientific discovery through agentic systems that autonomously conduct experiments and test hypotheses~\citep{lu2024ai, yamada2025ai, novikov2025alphaevolve, zhang2025darwin}. These frameworks leverage LLMs as sophisticated mutation operators, iteratively refining candidate solutions with successful variants propagating through successive generations. This methodology has proven effective across domains such as competitive programming \citep{li2022competition}, mathematical optimization \citep{romera2024mathematical}, and automated agentic design \citep{hu2024automated}.
However, current implementations face significant practical limitations. The primary challenge is substantial sample inefficiency as existing approaches typically require thousands of evaluations, making them computationally expensive and time-consuming. This inefficiency stems from naive exploration strategies that fail to effectively leverage accumulated knowledge from previous generations. Additionally, most leading systems remain closed-source, creating barriers to reproducibility and limiting community-driven improvements.
\ouralgo addresses these challenges through three key algorithmic innovations that work synergistically to enhance sample efficiency. Our adaptive parent and LLM sampling intelligently balances exploration of novel regions with exploitation of known high-quality areas. Next, our code proposal novelty rejection sampling ensures efficient program mutations. Finally, our bandit-based LLM ensemble selection strategy dynamically adapts to the evolving state of the sampled archive parents and inspiration programs.
Experimental validation across diverse domains demonstrates substantial improvements in both efficiency and solution quality, with \ouralgo achieving state-of-the-art results using orders of magnitude fewer evaluations than existing approaches. By releasing our complete implementation as open-source software, we aim to democratize access to advanced evolutionary discovery tools and enable broad community contributions. In summary:

\begin{enumerate}
    \item We introduce \ouralgo, an evolutionary framework that substantially improves sample efficiency through three key algorithmic innovations: a novel parent program sampling strategy, code novelty rejection-sampling, and adaptive performance-based LLM ensemble selection.
    \item We demonstrate \ouralgo's ability to innovate beyond human and LLM-generated solutions with comprehensive experimental validation across four distinct problem domains: mathematical optimization (circle packing), agentic design (AIME tasks), competitive programming (ALE-Bench), and LLM training design (mixture-of-expert load balancing loss). 
    \item We release \ouralgo as open-source software under the Apache 2.0 license, including implementation details, and an interactive visualization tool for monitoring the search process.
\end{enumerate}

\section{Related Work}

\textbf{Evolutionary Code Optimization with LLMs}. One particular flavor of test-time compute is evolutionary code optimization: the usage, mutation, and recombination of previously generated code to produce new samples. This approach has previously been used to optimize reward and preference objectives \citep{lu2024discovering, ma2023eureka}, mathematical science code \citep{romera2024mathematical}, and other applications \citep{lehman2022evolutionlargemodels, lange2024large, meyerson2023language, berman2025record, lange2025ai}. Through prompting, LLMs are used as recombination engines \citep{lange2023discovering_es, meyerson2023language}, and are capable of simulating crossover between diverse code snippets and the rationales that produced them. These types of program archive-building systems resemble a population-based LLM-guided tree search \citep{jiang2025aide, inoue2025wider}. Most closely related to our work are \textit{AlphaEvolve} \citep{novikov2025alphaevolve}, \textit{OpenEvolve} \citep{openevolve}, and \textit{LLM4AD} \citep{liu2024llm4ad}. We advance this line of work, demonstrating unprecedented sample efficiency with our combination of rejection-sampling, LLM prioritization, and online meta-scratchpad drafting.

\textbf{Open-Ended Agentic Discovery}. The integration of LLMs with open-ended evolutionary principles enables agentic systems capable of continuous innovation \citep{stanley2017open, zhang2025darwin}. Unlike traditional novelty search that relies on explicit diversity metrics \citep{lehman2008exploiting, lehman2011abandoning}, LLM agents leverage learned representations to generate creative solutions while maintaining semantic coherence \citep{omniepic, hu2024automated, novikov2025alphaevolve}. These agents construct evolutionary trees of programs where LLM-guided mutations connect related solutions across generations \citep{lehman2020surprising}. \ouralgo systematically combines stepping stones, suboptimal intermediate solutions that serve as building blocks for breakthrough innovations, by employing LLM agents to both generate mutations and evaluate program relationships, enabling successful patterns to rapidly propagate across search branches through recombination.

\section{Method}
\label{sec:method}

\textbf{Algorithm Overview.} \ouralgo's control-flow entails three main phases: 
\begin{enumerate}
\item \textit{Parent and inspiration sampling} from an archive of island program subpopulations. Importantly, we emphasize the trade-off between exploration and exploitation in parent program selection.
\item \textit{Program mutation} via LLM-guided code edit proposals. We utilize novelty rejection-sampling based on code embedding similarity and an LLM-as-a-novelty-judge assessment.
\item \textit{Program execution and world feedback} guiding the LLM ensemble selection probabilities and online meta-scratchpad drafting for documentation and knowledge diffusion.
\end{enumerate}

\subsection{Parent and inspiration sampling}
\textbf{Archive Maintenance, Island Populations \& Mutation Context Construction}. \ouralgo maintains a fixed-size archive of previously evaluated programs with fitness scores and meta information, implementing an elite size constraint. The mutation context incorporates a primary parent program alongside inspiration programs drawn from top-performing solutions and random archive samples, providing the LLM with diverse exemplars for creative recombination. We follow \citet{romera2024mathematical,novikov2025alphaevolve} and employ an island model approach with independent subpopulations seeded from the same initial program. The islands evolve in parallel to enhance diversity and prevent premature convergence. Island members can occasionally migrate between islands to diffuse knowledge across ``discovery substreams''. To protect the uniqueness of each island, we prevent the island-specific best-performing program from migrating~\citep{tanese1989distributed, romera2024mathematical}. Sampling occurs hierarchically: with the island ID first sampled uniformly from the archive, later used as the origin for both parent and inspirations. Afterwards, we sample random archive programs and the top-K performing programs to use them as context programs.

\begin{figure}[h!]
    \centering
    \includegraphics[width=0.925\textwidth]{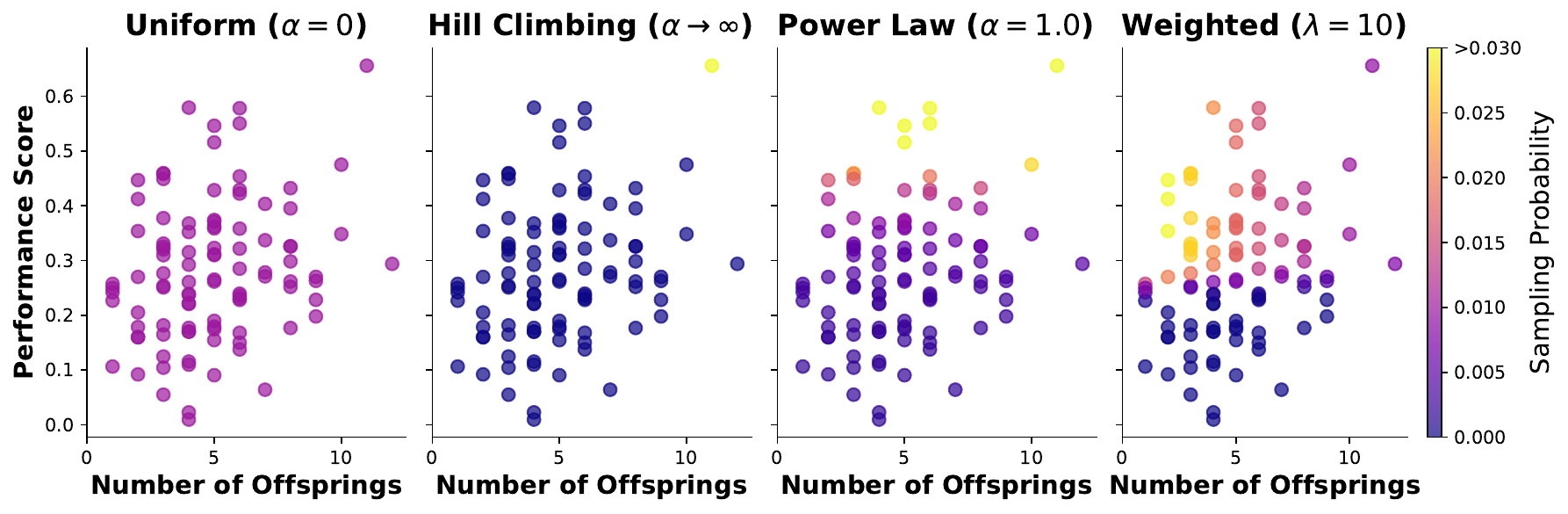}
    \caption{\textbf{\ouralgo Parent Sampling.} The strategies range from pure exploration (uniform sampling) to pure exploitation (hill-climbing) to a combination of performance and novelty.}
    \label{fig:parent_sampling}
\end{figure}

\textbf{Balancing Exploration \& Exploitation: Parent Program Selection}. Given an island subpopulation, \ouralgo implements multiple different parent sampling strategies that balance exploration and exploitation: 
First, we employ power law sampling where programs are ranked by fitness with ranks $r_i$ ($r_i = 1$ for the best program). The selection probability follows $p_i = \frac{r_i^{-\alpha}}{\sum_{j=1}^{n} r_j^{-\alpha}}$, where $\alpha$ controls exploitation intensity. Setting $\alpha = 0$ yields uniform sampling, while $\alpha \to \infty$ implements hill-climbing.
Inspired by \citet{zhang2025darwin}, we contrast this with weighted sampling, incorporating performance and novelty. Given programs with offspring count $N(P_i)$, we first compute the median fitness $\alpha_0 = \text{median}(\{F(P_1), F(P_2), ..., F(P_n)\})$. The performance component uses sigmoid scaling: $s_i = \sigma(\lambda \cdot (F(P_i) - \alpha_0))$ where $\sigma(x) = \frac{1}{1 + e^{-x}}$ and $\lambda$ controls selection pressure. The novelty component $h_i = \frac{1}{1 + N(P_i)}$ favors programs with fewer offspring. The final probability combines these: $p_i = \frac{w_i}{\sum_{j=1}^{n} w_j}$ where $w_i = s_i \cdot h_i$ balances performance and novelty. The strategies are illustrated in \Cref{fig:parent_sampling}.

\subsection{Program mutation and novelty assessment}

\textbf{LLM-Guided Program Mutations}.  To generate new programs, \ouralgo starts by sampling a specific LLM and a set of sampling parameters (e.g., temperature or reasoning budget) from a pre-specified pool. Our framework provides support for models from leading API providers, including GPT, Gemini, Claude, and DeepSeek~\citep{gpt4, moe_gemini25, claude3, guo2025deepseek}. After sampling a model, \ouralgo employs three distinct mutation approaches to foster diversity and creativity in the LLM-generated program variants: 
\begin{enumerate}
\item \textbf{Diff-Based Edits.} We implement diff edits using LLMs following the approach outlined in \citet{novikov2025alphaevolve}, utilizing SEARCH/REPLACE blocks for targeted modifications.
\item \textbf{Full Rewrites.} We enable full program rewrites to allow greater flexibility, programmatically ensuring that non-mutable blocks remain unchanged during the LLM rewrite process. 
\item \textbf{Crossover Mutation.} We leverage crossover mutations \citep{lehman2022evolutionlargemodels,lange2025ai} where an additional archive program is sampled and an LLM is prompted to combine programs.
\end{enumerate}

Following \citet{novikov2025alphaevolve}, we use text markers (\texttt{EVOLVE-BLOCK-START} \& \texttt{EVOLVE-BLOCK-END}) to ensure that immutable code is left unchanged during the LLM rewrite process. After obtaining a code change proposal, we enforce that the immutable code is not touched and resample a new proposal if a patch is invalid, providing parsing feedback using Reflexion \citep{shinn2024reflexion}. 

\begin{figure}[h!]
    \centering
    \includegraphics[width=0.925\textwidth]{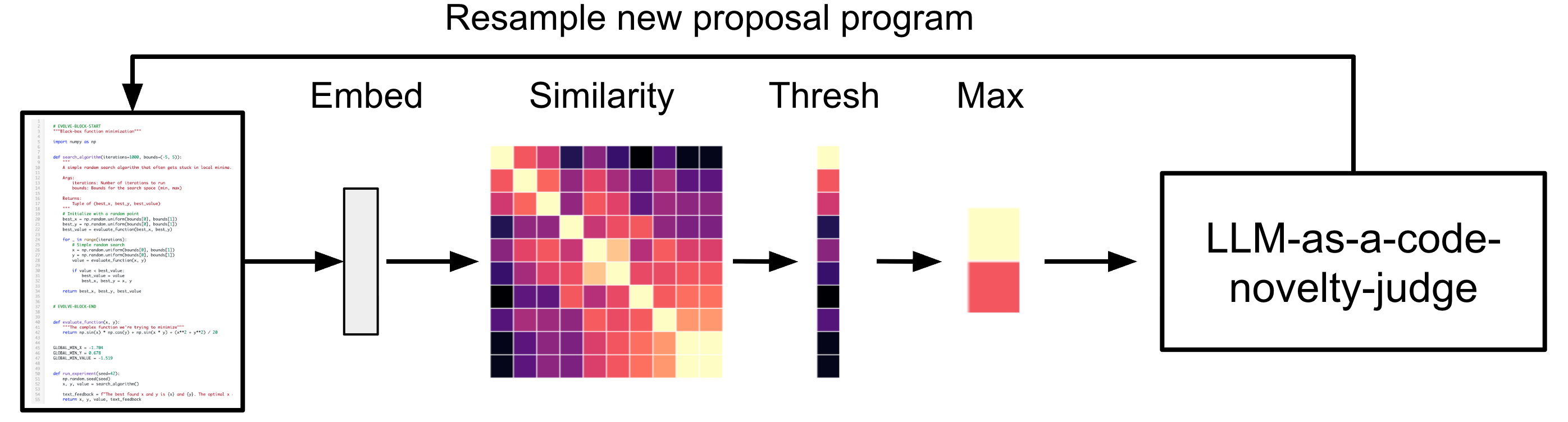}
    \caption{\textbf{\ouralgo Program Novelty Rejection Sampling.} \ouralgo embeds mutable code snippets, computes similarities across the archive; if the maximal score exceeds a threshold, another LLM is queried to assess whether the program is meaningfully novel.}
    \label{fig:novelty_rejection_sampling}
\end{figure}

\textbf{Program Diversity via Novelty Rejection Sampling}. To enhance the creativity of executed code proposals, we leverage a foundation model ensemble combined with temperature sampling. Additionally, we introduce \textit{code novelty rejection sampling} using an embedding model to embed mutable parts of the program code. Afterwards, we compute cosine similarity scores across the island subpopulation programs. If the maximal score exceeds a threshold (e.g., $\eta=0.95$), we query an LLM to further assess whether the program is meaningfully different. The approach is illustrated in \Cref{fig:novelty_rejection_sampling}.

\subsection{Execution and world feedback}

\textbf{Multi-Objective Optimization \& Textual Feedback}. After a program obtained with the above steps is executed, \ouralgo performs multi-objective assessment yielding both its scalar fitness value $r_i$ together with a set of exposed ``public metrics'' and textual feedback. \ouralgo then stores this full multi-objective assessment in the population archive to provide an informative context for future generations of language model mutations using a simple prompting format:

\begin{tcolorbox}[breakable,colback=orange!5!white, colframe=orange!80!black, title=Example of Diff Edit Prompt with Textual Feedback]
\tiny
\begin{Verbatim}[breaklines=true,breakanywhere=true]
# Current program
Here is the current program we are trying to improve (you will need to propose a modification to it below):
```{language}
{code_content}
```
Here are the performance metrics of the program:
{performance_metrics}{text_feedback_section}

# Instructions
...
# Task
...
IMPORTANT: Do not rewrite the entire program - focus on targeted improvements.
\end{Verbatim}
\end{tcolorbox}

\textbf{Adaptive LLM sampling evolution}. The performance of different LLMs to propose mutations can vary across problem domains and based on the current state of the sampled archive parents and inspiration programs. \ouralgo dynamically adapts to this non-stationarity by evolving the LLM sampling probability throughout at the end of each generation. Our approach is based on the UCB1 algorithm~\citep{ucb_sem}, associating each LLM with a visitation counter and an estimate of the expected score updated with the performance of its sampled mutations. We introduce changes tailored to the domain of LLM-driven discovery. In particular, rather than the absolute fitness of each mutation $r_i$, we update the LLM distribution using: $r_i^u = \exp\!\left(\max(r_i - r_i^b, 0)\right) - 1$, where $r_i^b$ is the baseline reward for program $i$ computed as the maximum between its parent program and the initial program in the database, ensuring each LLM is evaluated based on its relative improvement to account for the non-stationarity of the program archive. At the same time, the $\exp(\cdot)$ and $\max(\cdot, 0)$ operations help precisely promote LLMs able to come up with bold, high-risk, high-reward mutations,  over ``safer'' minor improvements. We use the tracked statistics over the observed rewards to normalize $r_i^u$ and ensure invariance to the fitness scale of each domain.

\begin{figure}[h!]
    \centering
    \includegraphics[width=0.9\textwidth]{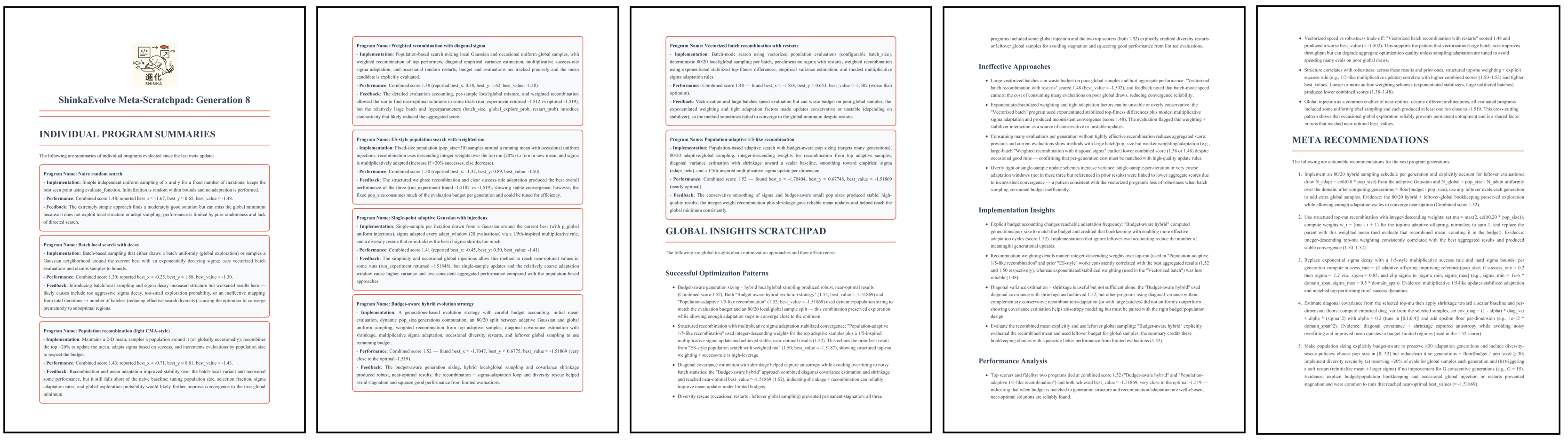}
    \caption{\textbf{A \ouralgo Meta-Scratchpad.} It consists of individual program summaries, global insights, and implementation recommendations, which are appended to the mutation prompt.}
    \label{fig:meta_scratchpad}
\end{figure}

\textbf{Meta-Scratchpad \& Online Refinement}. \ouralgo implements a meta-scratchpad system that periodically analyzes successful solutions to accelerate learning. Every $T$ generations, we summarize the recent program evaluations and identify common optimization strategies and design principles. The meta-agent synthesizes insights into actionable recommendations appended to the mutation prompt, providing high-level guidance from accumulated evolutionary experience. The approach is illustrated in \Cref{fig:meta_scratchpad}.

\newpage
\section{Results}
\label{sec:results}

\subsection{Circle Packing: Reproducing \& Improving AlphaEvolve Results}

\paragraph{Task Description.} The circle packing optimization problem requires placing 26 circles within a unit square such that the sum of their radii is maximized while ensuring no circles overlap and all circles remain fully contained within the square boundary. This constrained optimization challenge combines discrete placement decisions with continuous radius optimization, making it a complex benchmark for evolutionary algorithms. The problem exhibits multiple local optima and requires sophisticated search strategies to discover high-quality solutions, as naive approaches often converge to suboptimal configurations with poor space utilization.

\begin{figure}[h!]
    \centering
    \includegraphics[width=0.975\textwidth]{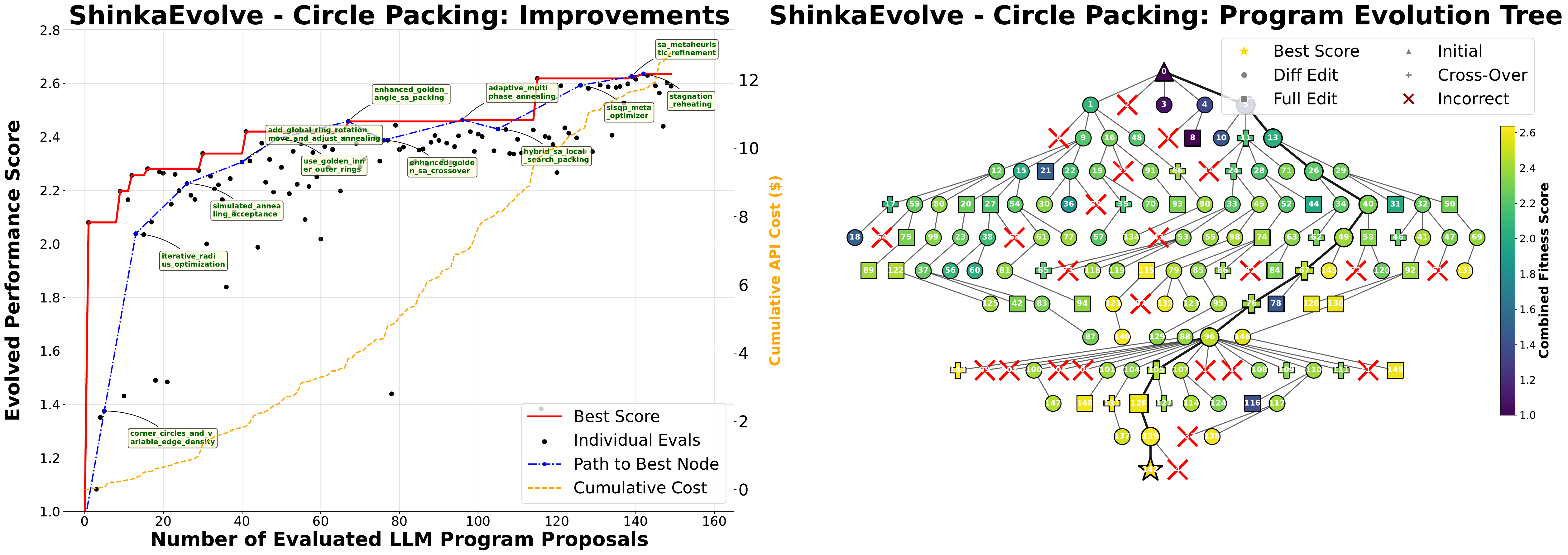}
    \caption{\textbf{\ouralgo on Circle Packing Task.} \textit{Left:} \ouralgo outperforms AlphaEvolve's solution within less than 150 program evaluations. \textit{Right:} \ouralgo's program evolution tree demonstrates the iterative composition of stepping stones into high-performing solutions.}
    \label{fig:results_circle_packing}
    \vspace{-15pt}
\end{figure}

\paragraph{\ouralgo's Discovery Dynamics.} \ouralgo was evaluated over 150 evolutionary generations, demonstrating remarkable sample efficiency compared to existing approaches that typically require thousands of evaluations (\Cref{fig:conceptual}).
\Cref{fig:results_circle_packing} (left) illustrates the improvement trajectory, exhibiting three distinct phases: an initial rapid improvement phase where the algorithm quickly discovers fundamental radii optimization strategies, a sustained exploration phase with incremental gains as more sophisticated techniques emerge (constraint-based optimization), and a final convergence phase where the best solutions are refined through restarts.
The tree structure in \Cref{fig:results_circle_packing} (right) reveals how successful innovations propagate through the population, with high-performing solutions (shown in green and yellow) serving as parents for subsequent generations.
Notably, the algorithm demonstrates sophisticated exploration patterns, with multiple evolutionary branches exploring different algorithmic approaches before converging toward the optimal solution path highlighted in black. We provide various ablation studies in \Cref{sec:ablations}.

\paragraph{\ouralgo's Discovered Solution.} The evolved algorithm (\Cref{appsec:circle_packing}) combines three key innovations: (1) a sophisticated initialization strategy that places circles in a structured golden-angle spiral pattern with strategic corner and edge positioning, (2) a hybrid optimization approach integrating SLSQP gradient-based refinement with simulated annealing for global exploration, and (3) intelligent perturbation mechanisms that alternate between local circle movements and global ring rotations to escape local optima. The discovered solution employs adaptive temperature scheduling with reheating strategies to prevent premature convergence, while maintaining feasibility through constraint-aware radius computation. This multi-level approach, from structured initialization through meta-heuristic exploration to gradient-based polishing, exemplifies how \ouralgo can discover sophisticated algorithmic compositions that outperform hand-designed baselines.

\newpage
\subsection{AIME: Evolving Agent Scaffolds for Math Reasoning}

\paragraph{Task Description.}
We evaluate \ouralgo on AIME 2024 \citep{AIME2024} mathematical reasoning problems, consisting of 30 challenging competition-level questions requiring sophisticated problem-solving strategies \citep{hu2024automated}.
The task involves evolving agent scaffold designs constrained to a maximum of 10 LLM queries per problem for computational efficiency.
Using \texttt{gpt-4.1-nano} as the base model, we discover scaffold designs for 75 generations, with each candidate evaluated across three independent runs on the complete question set.

\begin{figure}[h!]
    \centering
    \includegraphics[width=0.975\textwidth]{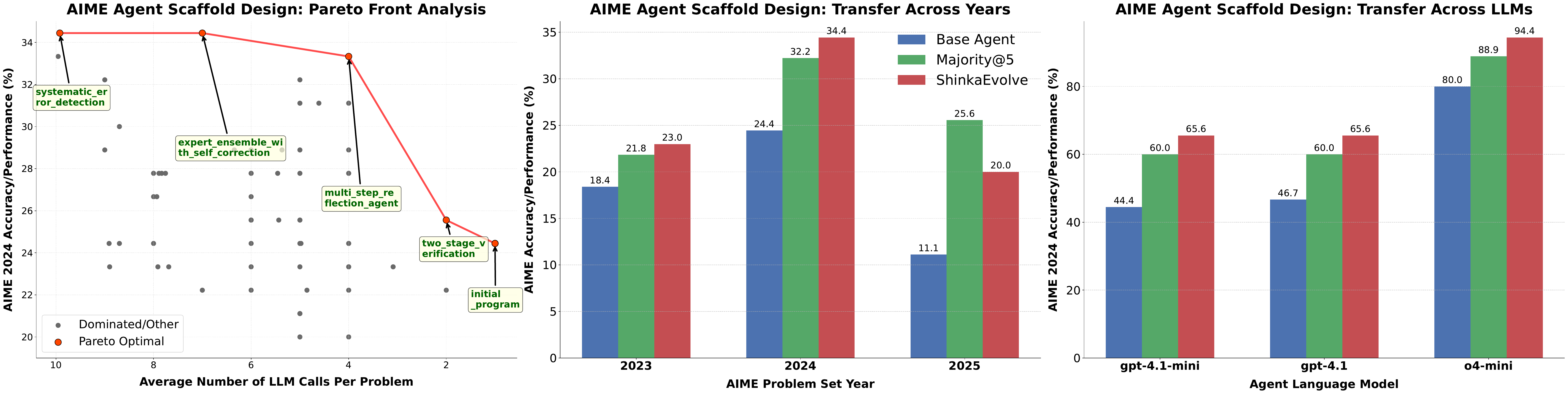}
    \caption{\textbf{\ouralgo for Agent Scaffold Design.} \textit{Left:} \ouralgo discovers a Pareto frontier between performance and LLM query budget. \textit{Middle:} The discovered scaffold generalizes to unseen AIME problems. \textit{Right:} The scaffold improves performance regardless of the underlying LLM.}
    \label{fig:results_aime}
    \vspace{-15pt}
\end{figure}

\paragraph{\ouralgo's Discovery Dynamics.}
The evolutionary process systematically explores prompting strategies, ensemble methods, and verification techniques to identify optimal agent architectures.
\ouralgo discovers scaffold designs that significantly outperform hand-designed baselines, including simple single-query agents and sophisticated majority-voting approaches.
The search reveals a Pareto frontier between efficiency and performance (\Cref{fig:results_aime}, left), with 7 LLM queries yielding maximum performance while an alternative scaffold achieves comparable results using the full 10-query budget.
Generalization experiments reveal important insights into the scaffold's robustness.
Evaluating on 2023 and 2025 AIME problems shows different transfer patterns (\Cref{fig:results_aime}, middle): smaller improvements on 2023 problems suggest potential saturation due to training data contamination, while larger gains on 2025 problems indicate successful generalization to recent, unseen challenges.
Cross-LLM model transfer experiments validate robustness, with successful adaptation to \texttt{gpt-4.1-mini}, \texttt{gpt-4.1}, and \texttt{o4-mini} demonstrating that discovered architectures capture generalizable strategies rather than model-specific optimizations (\Cref{fig:results_aime}, right).

\paragraph{\ouralgo's Discovered Solution.}
The evolved agent implements a three-stage architecture leveraging diverse expert personas, critical peer review, and synthesis mechanisms.
Three specialized experts generate independent solutions using distinct approaches: a meticulous step-by-step reasoner, an intuitive pattern-recognition specialist, and an algorithmic computer science-oriented mathematician, each operating at 0.7 temperature to balance creativity with reliability.
The second stage introduces critical peer review, where each solution undergoes rigorous scrutiny from a skeptical reviewer at low temperature (0.1).
The reviewer validates pattern-based reasoning by testing patterns on multiple examples, identifies logical flaws, and provides corrections when necessary, significantly improving solution quality.
The final synthesis stage employs an editor-in-chief persona operating at zero temperature to analyze all solutions and critiques, identify the most reliable approach, and construct a canonical solution.
Robust fallback mechanisms resort to majority voting among reviewed solutions, then original solutions, ensuring reliable output when components fail.
This architecture effectively utilizes 7 LLM calls (3 generation + 3 review + 1 synthesis) within the 10-call constraint.
The complete discovered agent scaffold can be found in \Cref{appsec:aime_agent}.

\newpage
\subsection{ALE-Bench: Evolving Programs for Combinatorial Optimization}

\paragraph{Task Description.} We apply \ouralgo to the ALE-Bench LITE \citep{imajuku2025ale} benchmark, a collection of 10 competitive programming contests hosted by AtCoder and designed to test the performance of LLMs on heuristic problems. Here, we explore whether \ouralgo can successfully improve high-performing solutions discovered by LLMs. We leverage the best programming solution discovered by ALE-Agent \citep{imajuku2025ale} as an initial program for each problem and apply \ouralgo to improve on top of it. We run \ouralgo for 50 generations, leveraging the score calculated on the public test set as the fitness function. Afterwards, we submit the best solution to the private test set and report the score.

\begin{figure}[h!]
    \centering
    \includegraphics[width=0.975\textwidth]{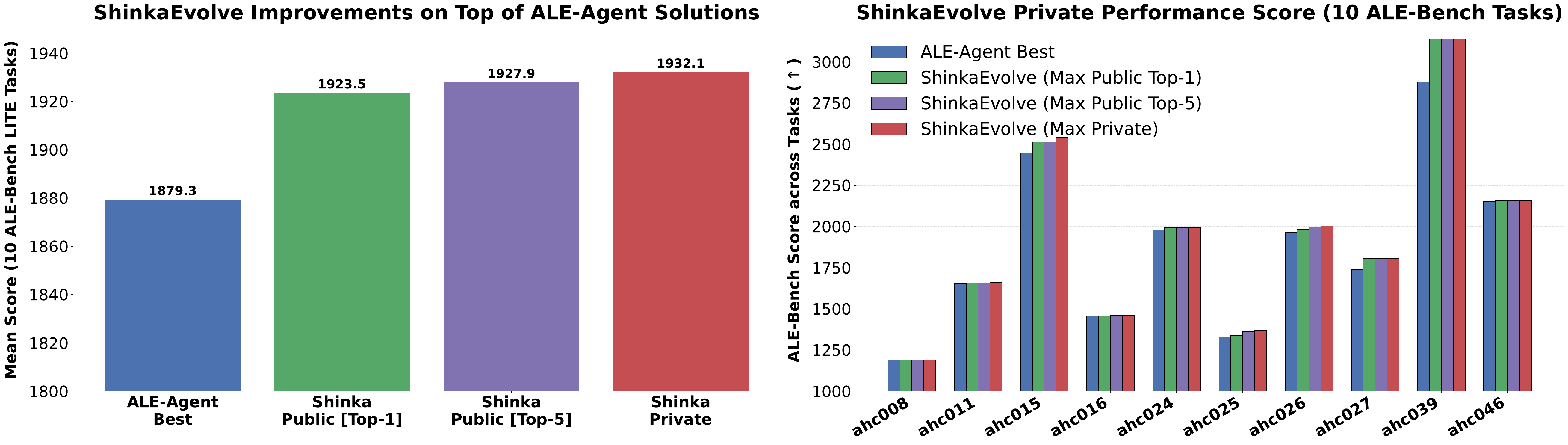}
    \caption{\textbf{\ouralgo for Improving ALE-Bench solutions.} \textit{Left:} \ouralgo improves the solutions discovered by ALE-Agent by $\sim2.3\%$. \textit{Right:} On one task, \texttt{ahc039}, the solution improved from 5th to 2nd place submission on the AtCoder leaderboard if it had participated in the contest.}
    \label{fig:results_alebench}
    \vspace{-15pt}
\end{figure}

\paragraph{\ouralgo's Discovery Dynamics.} \ouralgo is able to improve the solutions discovered by ALE-Agent by approximately 2.3\% across the 10 tasks on average (\Cref{fig:results_alebench}). Furthermore, on one task, \texttt{ahc039}, the combination of \ouralgo with ALE-Agent resulted in the second place submission on the \href{https://atcoder.jp/contests/ahc039/submissions/68865341}{AtCoder leaderboard} if they had participated. While these improvements resulted from detailed implementation improvements, we observe that the proposed changes by \ouralgo remained algorithmically close to the original ALE-Agent's initialization solution.

\paragraph{\ouralgo's Discovered Solution.}
We focus on two tasks to illustrate the discovered improvements of \ouralgo, \texttt{ahc039} and \texttt{ahc025}.
The objective of \texttt{ahc039} is to find an optimal, axis-aligned polygon to maximize the number of mackerels it contains minus the number of sardines, subject to given constraints. The base solution by ALE-Agent applies simulated annealing with kd-tree data structure (5th, 2880 performance). \ouralgo further improved the solution (2nd, 3140 performance) by introducing modifications such as caching the validation process and enhancing neighborhood operators. For the caching, the kd-tree was augmented to cache subtree statistics, including bounding boxes and fish counts, at each node. For the neighborhood operators, a novel ``targeted edge move'' was introduced, which heuristically identifies a misclassified fish (e.g., a mackerel outside the polygon) and greedily moves the nearest edge to correct its state. These changes strengthened the directionality of the search.
For \texttt{ahc025}, the task is to use a balance scale to compare the total weights of any two subsets of items, aiming, after a fixed number of weighings, to partition the items into groups with as equal total weights as possible. \ouralgo improved the ALE-Agent's simulated annealing baseline by introducing faster caching, refining fallback weight estimation, and ultimately replacing simulated annealing with a more focused optimization combining greedy moves and targeted local search. Comparison with top human solutions suggests that for many tasks, there is ample room for improvement. Furthermore, often times \ouralgo tended to explore modifications staying close to the ALE-Agent's solution. This indicates the potential of overfitting to the initialization solution.

\newpage
\subsection{LLM Training: Evolving Losses for Balanced and Effective Experts}
\paragraph{Task Description.} The Mixture-of-Expert (MoE) architecture \citep{moe_sem_1, moe_sem_2, moe_mod_1_2exp, moe_mod_2_switch_1exp} has been a critical advancement, ubiquitous amongst modern open and closed-source flagship models~\citep{moe_gemini15, guo2025deepseek, llama4, moe_qwen3,  moe_gemini25}. The basic idea is simple: replace traditional large feed-forward residual blocks with ensembles of efficient smaller modules (the ``experts'') that can each specialize in distinct problem domains~\citep{moe_mod_2_switch_1exp}. For each MoE layer and token, only the outputs of the top-K experts selected by a router classifier are computed, effectively splitting the computation and making both training and inference cheaper and faster. However, due to the non-differentiability of the top-K expert selection operation, it is critical to provide the router with an auxiliary load balancing loss (LBL), which serves to avoid early collapse toward uneven expert distribution of the token load. We deploy \ouralgo precisely to tackle this open architectural design challenge, which has been one core focus driving recent MoE advancements ~\citep{moe_sem_2, moe_mod_2_switch_1exp, moe_lbl_evo1, moe_lbl_evo2_zloss, moe_lbl_evo3, moe_deepseek_moe, moe_deepseek_demon, moe_similar_olmoe}: Devising an effective load balancing loss to incentivize efficiency and specialization, without hindering the model's expressivity.

\begin{figure}[h!]
    \centering
    \includegraphics[width=0.975\textwidth]{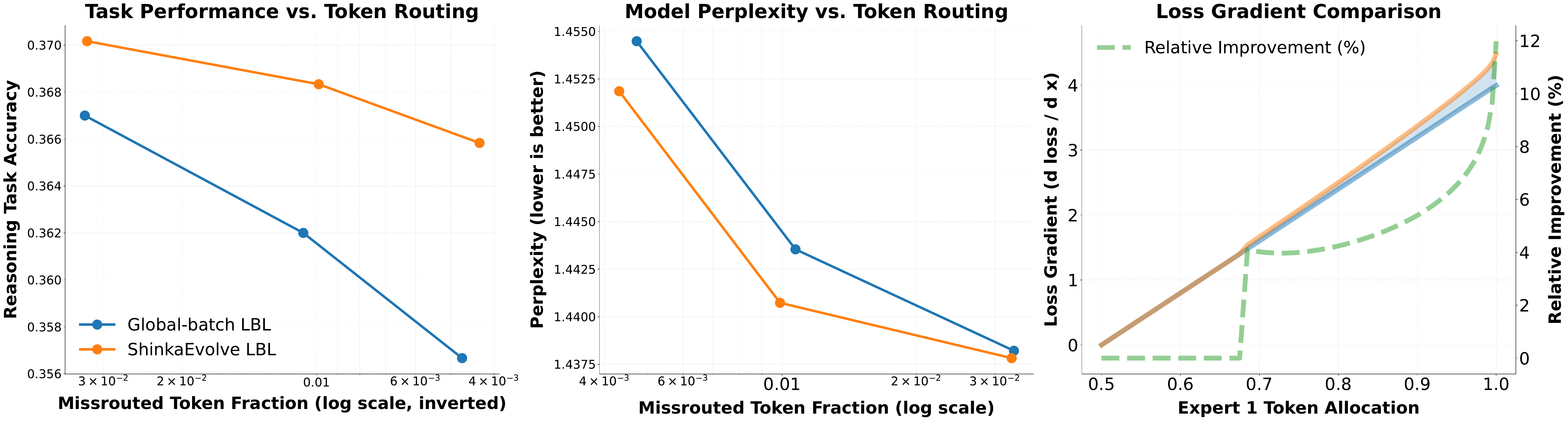}
    \caption{\textbf{\ouralgo for discovering Mixture-of-Experts Load Balancing Loss Functions.} \textit{Left:} Downstream task performance across seven benchmarks. \textit{Middle:} Final perplexity across different missroute fractions. \textit{Right:} Load imbalance gradient as a function of the token allocation.}
    \label{fig:results_moe}
    \vspace{-15pt}
\end{figure}

\paragraph{\ouralgo's Discovery Dynamics.} We ground the problem of LBL design by pretraining a MoE model with 556M parameters, $N_E=64$ total experts of which only $K=8$ active for any given token. This results in only 82M parameters sparsely activated in each forward pass, excluding the token embeddings. We train this small model on over 2B tokens from fineweb~\citep{fineweb} by minimizing the MoE loss function adding the LBL, weighted by $\lambda=0.01$, to the model's cross-entropy loss (CE). The fitness function of each program then measures a simple objective:  minimize the sum of the final CE together with the model's ``load imbalance'' as measured by the L1 deviation from a uniform distribution of tokens between the MoE experts. Given the cost of pretraining, we run \ouralgo for only 30 iterations.
We evaluate the generality of \ouralgo's best-performing solutions by training a much larger MoE with 2.7B parameters on slightly under 30B fineweb tokens across three LBL coefficients  $\lambda\in{0.001,0.01, 0.1}$, yielding different levels of regularization. We then compare with the ``global-batch LBL'' used to train some of the most popular open LLMs~\citep{moe_qwen3}, in terms of final perplexity (Figure~\ref{fig:results_moe}, left) and downstream task performance (Figure~\ref{fig:results_moe}, center) as evaluated across seven different benchmarks \citep{moe_bench_1_commonsenseqa, moe_bench_2_hellaswag, moe_bench_3_openbook_qa, moe_bench_4_piqa, moe_bench_5_siqa, moe_bench_6_winogrande, moe_bench_7_arc}. We provide our results below as a function of load imbalance, showing that \ouralgo's new loss achieves a consistent edge across our training configurations, growing larger with the value of the $\lambda$ coefficient.
\paragraph{\ouralgo's Discovered Solution.} The discovered LBL is a new twist on the established global-batch LBL, which was used for seeding the evolutionary search. \ouralgo complements this popular LBL with a new term, specifically targeted toward regularizing the MoE layers with individual under-specialized experts. Concretely, let $f_{\ell, i}$ and $P_{\ell, i}$ correspond to the selection frequency and the average router probabilities for each expert $i$ located in layer $\ell$. \ouralgo's LBL uses a normalized complement to the entropy in each layer $s(P_\ell)=0.5 + \Bigl(1-\frac{H(P_\ell)}{\log N_E}\Bigr)$ and a minimum usage threshold target $\tau = 0.064/{N_E}$ to compute:
\begin{equation}
\label{eq:discovered_lbl_colored}
\begin{aligned}
L_{\mathrm{LBL}}
&=
\underbrace{%
\color{lblA} N_E \cdot \frac{1}{L}
\sum_{\ell=1}^{L}\sum_{i=1}^{N_E} f_{\ell,i}\,P_{\ell,i}
}_{\text{Global-batch LBL}}+
\underbrace{%
\color{lblB} \frac{0.1}{L}\sum_{\ell=1}^{L}
s(P_\ell)\,\sum_{i=1}^{N_E}\max\!\bigl(0,\tau-f_{\ell,i}\bigr)
}_{\text{\ouralgo new regularization}} \, .
\end{aligned}
\end{equation}
The effects of \ouralgo's new regularization term can be visualized through its induced gradients acting on the router's token allocation in a simplified two-expert scenario (Figure~\ref{fig:results_moe}, right). Intuitively, this term softly affects the MoE router of any layer, with experts getting allocated a fraction of tokens less than $\tau$. The multiplier $s(P_\ell)$ makes this push stronger when the layer's routing entropy $H(P_\ell)$ is low and the router is concentrating on fewer dominating experts. This closes a potential blind spot of the global-batch LBL: the dot product $f\!\cdot\!P$ can look ``balanced'' even if few experts are barely touched. Thus, \ouralgo's new term can be seen as a safety net that adaptively activates and vanishes once an expert crosses the floor, providing dead experts and avoiding over-regularizing well-balanced layers.


\section{Ablations \& Analysis} 
\label{sec:ablations}
\begin{figure}[h!]
    \centering
    \includegraphics[width=0.975\textwidth]{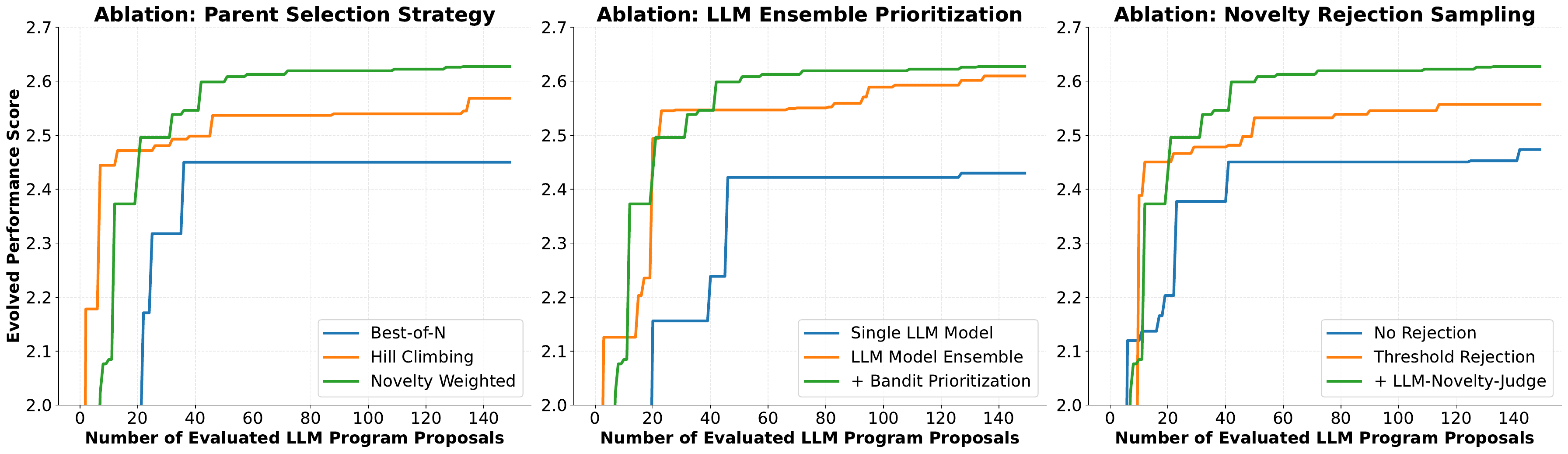}
    \caption{\textbf{\ouralgo Method Ablation Studies on Circle Packing.} \textit{Left:} Weighted parent sampling outperforms random search and hill climbing. \textit{Middle:} Bandit-based LLM ensembling slightly improves the performance over a fixed uniform ensemble distribution. \textit{Right:} Embedding-based rejection sampling with LLM as a novelty judge strongly outperforms no rejection sampling.}
    \label{fig:ablations}
    \vspace{-15pt}
\end{figure}

\paragraph{Impact of Parent Selection Strategies.}
To understand the importance of parent selection, we compare different strategies for choosing which programs to evolve. The \textit{Best-of-N} baseline ignores the evolutionary history, always using the initial program as parent without feedback. In contrast, \textit{Hill Climbing} represents a greedy approach that consistently selects the highest-performing program as the parent for subsequent mutations. Our proposed \textit{Weighted Sampling} strategy balances exploration and exploitation by probabilistically selecting parents based on their fitness and number of offspring.

\textit{Takeaways.} Weighted sampling consistently outperforms both random search and hill climbing across all tasks. Hill climbing shows strong initial performance but plateaus quickly, while weighted sampling maintains steady improvement throughout evolution. Random search demonstrates the poorest convergence, highlighting the importance of leveraging fitness-based parent selection.

\paragraph{Impact of LLM Ensembling and Prioritization.}
Evolutionary agents can benefit from diverse coding capabilities by leveraging multiple LLMs. We investigate this hypothesis by comparing a \textit{Single LLM} baseline (GPT-5-nano) against ensemble approaches. The \textit{Fixed LLM Ensemble} provides diversity by sampling uniformly from a predetermined set of models, while our \textit{Bandit-Based LLM Ensemble} adaptively learns which models contribute most effectively to fitness improvements, balancing exploration of underutilized models with exploitation of high-performing ones.

\textit{Takeaways.} The bandit-based LLM ensemble significantly outperforms both single LLM and fixed ensemble approaches. While the fixed ensemble shows moderate improvements over single LLM usage, the adaptive bandit strategy achieves the highest performance by dynamically prioritizing more effective models based on their contribution to fitness improvements.

\paragraph{Impact of Code Embedding-Based Rejection Sampling.}
Similar code variants can waste computational resources without advancing the search frontier. To address this challenge, we examine different novelty filtering mechanisms. The \textit{No Rejection Sampling} baseline accepts any LLM proposal, potentially allowing near-duplicate programs to proliferate. Our \textit{Embedding-Based Rejection Sampling} approach leverages text embeddings to identify and reject proposals with similarity scores exceeding 0.95. We also explore an \textit{Additional LLM-as-a-novelty-judge} variant that supplements embedding-based filtering with explicit LLM assessment of program novelty.

\textit{Takeaways.} Code embedding-based rejection sampling provides substantial performance gains over no rejection sampling by preventing redundant mutations. The additional LLM-as-a-novelty-judge offers marginal improvements, suggesting that embedding similarity is already an effective proxy for novelty assessment without requiring additional computational overhead.



\section{Discussion}

\paragraph{Summary.}
This work introduces \ouralgo, an evolutionary framework addressing critical limitations in LLM-driven scientific discovery through improved sample efficiency and open-source accessibility.
\ouralgo achieves state-of-the-art results across four domains: circle packing with 150 evaluations (orders of magnitude improvement), sophisticated AIME reasoning scaffolds, ALE-Bench algorithmic improvements, and novel mixture-of-expert load balancing.
\paragraph{Limitations.}
Our implementation uses fixed configurations with limited automatic control over exploration-exploitation balance, which may vary across domains.
Task specification requires manual human expertise for objective functions and evaluation.
The framework is constrained to problems with well-defined numerical objectives, limiting its applicability to diverse evaluation domains.
\paragraph{Future Directions.}
Automated task specification through LLM task generation could enable greater autonomy and unlock applications in unexplored domains.
Transitioning to true open-endedness, where systems generate their own objectives, represents a compelling frontier.
Self-referential refinement and online meta-learning offer opportunities for continuously improving discovery.
\paragraph{Broader Impact \& Ethical Considerations.}
\ouralgo's open-source release further democratizes advanced evolutionary optimization, making it accessible to researchers and practitioners previously lacking access to proprietary systems.
The framework's exceptional sample efficiency reduces computational barriers for resource-constrained environments.
However, API costs from large-scale LLM usage could create economic barriers, potentially constraining democratization goals.

\newpage
\section*{Acknowledgments}

We thank David Ha, Takuya Akiba, Taishi Nakamura, Luca Grilloti, Yutaro Yamada, and the rest of the Sakana AI team for helpful discussions throughout the project.

\section*{Author Contribution List}
\label{sec:author_contribution_list}

\paragraph{Robert Tjarko Lange} Initiated and led the project, designed the core \ouralgo codebase, and implemented as well as collected results for Circle Packing, AIME, and ALE-Bench. Wrote the manuscript.
\paragraph{Yuki Imajuku} Helped setting up the ALE-Bench infrastructure, advised on the ALE-Bench results and supported writing the manuscript section on ALE-Bench.
\paragraph{Edoardo Cetin} Was involved in design discussions for \ouralgo and came up as well as implemented the adaptive LLM sampling method and the Hydra configuration. He implemented andcollected results for LLM training and Mixture-of-Experts evolution. Co-wrote the manuscript.

\newpage
\bibliography{references}
\bibliographystyle{plainnat}

\clearpage
\appendix

\etocdepthtag.toc{mtappendix}
\etocsettagdepth{mtchapter}{none}
\etocsettagdepth{mtappendix}{subsection}
\renewcommand{\contentsname}{Appendix}
\tableofcontents

\newpage
\section{Shinka Implementation Details}
\label{appsec:shinka_code}

\begin{itemize}
  \item \ouralgo uses a queue based implementation where LLMs generate program proposals sequentially. Afterwards, they are added to a job evaluation queue. Each proposal is based on all jobs that have completed so far and are stored in the database.
  \item Throughout development, we experimented with a fully asynchronous implementation that leverages both a job and a proposal queue. This allows for higher throughput but introduces a degree of "off-archiveness" in the sense that new code proposals are generated in advance and not based on all the previously submitted jobs. Furthermore, jobs from faster to query models will be executed earlier since their proposal jobs will be processed earlier. Many open research questions remain regarding the optimal trade-off between throughput, sample efficiency, and off-archiveness.
  \item Below we provide an overview of the Python API. It roughly adopts the high-level interface of \texttt{OpenEvolve} \citep{openevolve}:
\end{itemize}

\begin{minipage}{\textwidth}
\begin{lstlisting}[language=Python, basicstyle=\ttfamily\tiny, caption={Minimal \ouralgo configuration and usage example.}, label={lst:shinka_usage}]
from shinka.core import EvolutionRunner, EvolutionConfig
from shinka.database import DatabaseConfig
from shinka.launch import LocalJobConfig

# Minimal config - only specify what's required
job_config = LocalJobConfig(eval_program_path="evaluate.py")
db_config = DatabaseConfig()
evo_config = EvolutionConfig(init_program_path="initial.py",)

# Run evolution with defaults
runner = EvolutionRunner(
    evo_config=evo_config,
    job_config=job_config,
    db_config=db_config,
)
runner.run()
\end{lstlisting}
\end{minipage}

\vspace{0.5em}

\begin{minipage}{0.45\textwidth}
\textbf{\texttt{evaluate.py} - Evaluation Script}

\begin{lstlisting}[language=Python, basicstyle=\ttfamily\tiny]
from shinka.core import run_shinka_eval

def main(program_path: str,
         results_dir: str):
    metrics, correct, err = run_shinka_eval(
        program_path=program_path,
        results_dir=results_dir,
        experiment_fn_name="run_experiment",
        num_runs=3, # Multi-evals to aggreg.
        get_experiment_kwargs=get_kwargs,
        aggregate_metrics_fn=aggregate_fn,
        validate_fn=validate_fn,  # Optional
    )

def get_kwargs(run_idx: int) -> dict:
    return {"param1": "value", "param2": 42}

def aggregate_fn(results: list) -> dict:
    score = results[0]
    text = results[1]
    return {
        "combined_score": float(score),
        "public": {...},  # shinka-visible
        "private": {...},  # shinka-invisible
        "extra_data": {...},  # store as pkl
        "text_feedback": text,  # str fb
    }

if __name__ == "__main__":
    # argparse program path & dir
    main(program_path, results_dir)
\end{lstlisting}
\end{minipage}
\hfill
\begin{minipage}{0.45\textwidth}
\textbf{\texttt{initial.py} - Starting Solution}

\begin{lstlisting}[language=Python, basicstyle=\ttfamily\tiny]
# EVOLVE-BLOCK-START
def advanced_algo():
    # This will be evolved
    return solution
# EVOLVE-BLOCK-END

def run_experiment(**kwargs):
    """Main called by evaluator"""
    result = solve_problem(kwargs)
    return result

def solve_problem(params):
    solution = advanced_algo()
    return solution
\end{lstlisting}
\end{minipage}

\clearpage
\section{Task Implementation Details}
\label{appsec:implememtation}
\subsection{Circle Packing Problem}

\textbf{Detailed Task Description.} The circle packing task requires placing 26 circles within a unit square such that the sum of their radii is maximized while ensuring no circles overlap and all circles remain fully contained within the square boundary.

\begin{minipage}{0.55\textwidth}
\textbf{Verification Methodology with Slack.} For the main 
\ouralgo run presented in the paper, we employed the 
verification script provided by \texttt{OpenEvolve} \citep
{openevolve}, which allows for $1 \times 10^{-6}$ numerical 
slack. To ensure the robustness of our results, we 
additionally validated our solutions using \texttt
{AlphaEvolve}'s \citep{novikov2025alphaevolve} exact 
verification code. We found that our discovered solution can 
be made trivially exact by reducing each circle's radius by 
$1 \times 10^{-8}$, demonstrating the high precision of our 
approach. The adjustment from the relaxed to exact 
formulation reduces the sum of radii for our discovered 
solution by a negligible amount, from 2.635983099011548 to 2.
6359828390115476, representing a relative change of less than 
$10^{-6}$. 
\end{minipage}
\hfill
\begin{minipage}{0.35\textwidth}
    \centering
    \includegraphics[height=5.5cm]{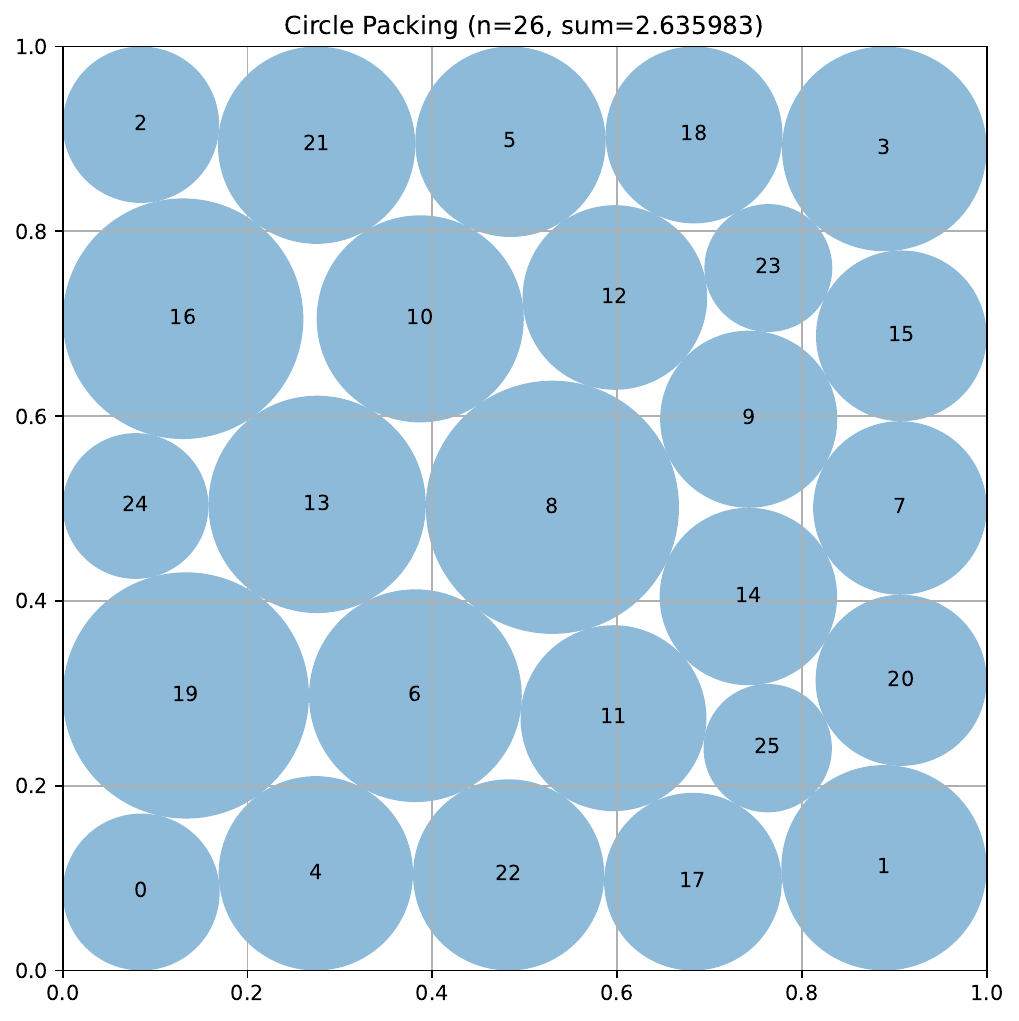}
    \captionof{figure}{Discovered Circle Packing solution by \ouralgo.}
    \label{fig:circle_packing_async_solution}
\end{minipage}

\begin{figure}[h]
\centering
\includegraphics[width=0.75\textwidth]{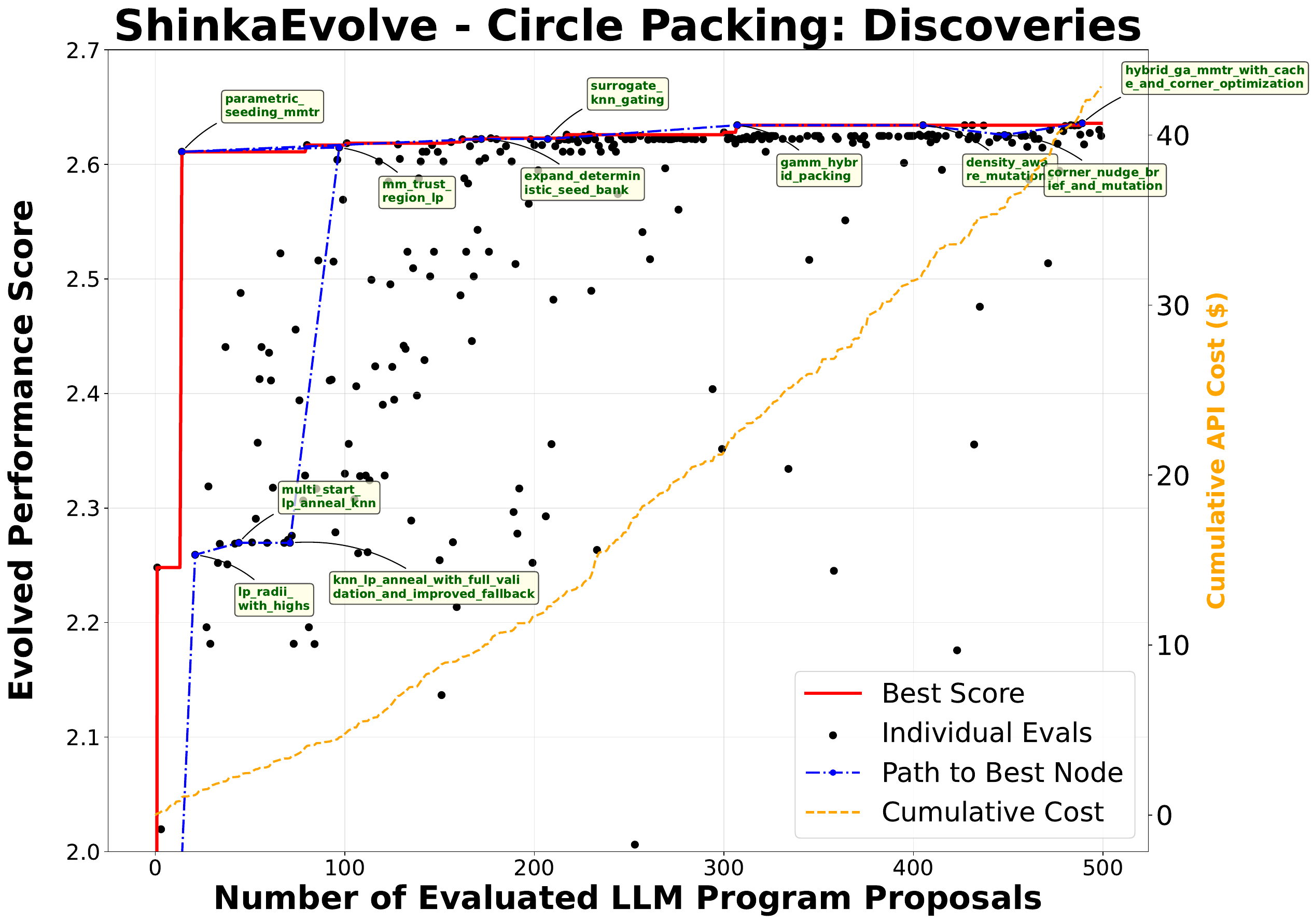}
\caption{Circle packing asynchronous evolution results for exact circle packing verification showing convergence behavior and solution quality over time.}
\label{fig:circle_packing_async}
\end{figure}

\textbf{Verification Methodology with Exact Constraint.} Additionally, we replicated the solution using the exact verification code from \texttt{AlphaEvolve} \Cref{fig:circle_packing_async} with a score of 2.63597770931127. The discovery of the solution requires more samples to be evaluated. This illustrates an important principle: surrogate relaxed tasks can be effectively used during evolution and subsequently post-processed to discover exact state-of-the-art solutions.

\textbf{Baseline Comparisons.} Our performance benchmarks are established against solutions from three primary sources. The AlphaEvolve sum of radii is taken from their paper \citep{novikov2025alphaevolve}. The \texttt{OpenEvolve} baseline scores are derived from their \href{https://github.com/codelion/openevolve/tree/main/examples/circle_packing}{official implementation and examples} available in their repository. Additionally, we compare against \texttt{LLM4AD} results, specifically their \href{https://github.com/Optima-CityU/LLM4AD/blob/main/example/circle_packing/circle_packing_result.ipynb}{circle packing implementations} and \href{https://github.com/Optima-CityU/LLM4AD/blob/main/example/circle_packing/EoH_settings%26logs/run_eoh.py}{Evolution of Heuristics (EoH) experimental configurations}. These baselines provide comprehensive coverage of existing automated algorithm design approaches, enabling fair and thorough performance evaluation of our method.

\textbf{\ouralgo's Hyperparameter Configuration.}

\begin{table}[h]
\centering
\small
\begin{tabular}{lclc}
\toprule
\textbf{Parameter} & \textbf{Value} & \textbf{Parameter} & \textbf{Value} \\
\midrule
\multicolumn{4}{l}{\textbf{Database configuration}} \\
\midrule
Archive size & 40 & Elite selection ratio & 0.3 \\
Archive inspirations & 4 & Top-$k$ inspirations & 2 \\
Migration interval & 10 & Migration rate & 0.0 \\
Island elitism & true & Parent selection strategy & weighted \\
Parent selection $\lambda$ & 10.0 & Number of islands & 2 \\
\midrule
\multicolumn{4}{l}{\textbf{Evolution configuration}} \\
\midrule
Patch types & [diff, full, cross] & Patch type probs & [0.45, 0.45, 0.1] \\
Generations & 150 & Max parallel jobs & 5 \\
Max patch resamples & 3 & Max patch attempts & 3 \\
Meta recommendation interval & 10 & Max meta recommendations & 5 \\
Embedding model & text-embedding-3-small & Max novelty attempts & None \\
Code embed sim threshold & 0.95 & Problem implementation & Python \\
LLM dynamic selection & ucb1 & Exploration coefficient & 1.0 \\
\midrule
\multicolumn{4}{l}{\textbf{LLM models}} \\
\midrule
gemini-2.5-pro & $\times$ & gemini-2.5-flash & $\times$ \\
claude-sonnet-4 & $\checkmark$ & o4-mini & $\checkmark$ \\
gpt-5 & $\times$ & gpt-4.1-nano & $\checkmark$ \\
gpt-4.1 & $\checkmark$ & gpt-4.1-mini & $\checkmark$ \\
\midrule
\multicolumn{4}{l}{\textbf{LLM settings}} \\
\midrule
Temperatures & [0.0, 0.5, 1.0] & Max tokens & 16{,}384 \\
Meta models & [gpt-5-nano] & Meta temperatures & [0.0] \\
Novelty models & [gpt-5-nano] & Novelty temperatures & [0.0] \\
\bottomrule
\end{tabular}
\caption{\ouralgo{} hyperparameter configuration for the Circle Packing task.}
\label{tab:hyperparams_circle}
\end{table}

\clearpage
\subsection{AIME Math Reasoning Agentic Harness}

\begin{minipage}{0.45\textwidth}
  \paragraph{Detailed Task Description.} For the agent scaffold design task, we evaluate \ouralgo on AIME 2024 mathematical reasoning problems, consisting of 30 challenging competition-level questions requiring sophisticated problem-solving strategies \citep{AIME2024}. We limit the maximum number of LLM queries per problem to 10 for computational and cost efficiency. Using \texttt{gpt-4.1-nano} as the base model, we evolve scaffold designs over 75 generations. Additionally and to combat stochasticity in LLM queries, we evaluated each candidate evaluated across three independent runs on the complete question set. After evolution, we evaluate the discovered scaffold designs on 2023 and 2025 AIME problems \citep{AIME2023, AIME2025} to assess generalization as well as robustness to different base agent language models.
  \end{minipage}
  \hfill
  \begin{minipage}{0.45\textwidth}
      \centering
      \includegraphics[height=5.5cm]{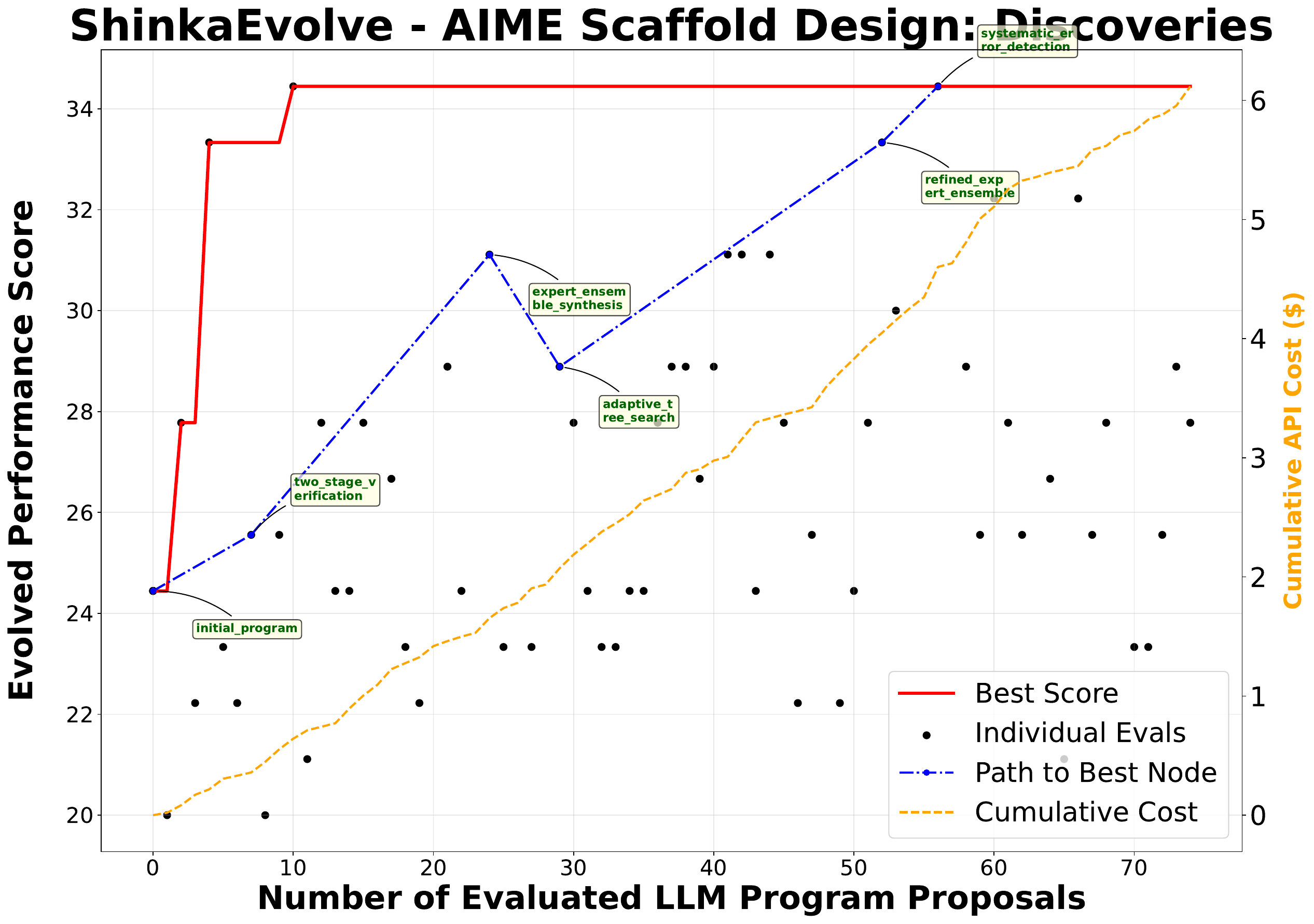}
      \captionof{figure}{\ouralgo's Discovery Trajectory for Math Agent Scaffold Design.}
      \label{fig:aime_solution}
  \end{minipage}

\textbf{\ouralgo's Hyperparameter Configuration.}

\begin{table}[h]
\centering
\small
\begin{tabular}{lclc}
\toprule
\textbf{Parameter} & \textbf{Value} & \textbf{Parameter} & \textbf{Value} \\
\midrule
\multicolumn{4}{l}{\textbf{Database configuration}} \\
\midrule
Archive size & 40 & Elite selection ratio & 0.3 \\
Archive inspirations & 4 & Top-$k$ inspirations & 2 \\
Migration interval & 10 & Migration rate & 0.1 \\
Island elitism & true & Parent selection strategy & weighted \\
Parent selection $\lambda$ & 10.0 & Number of islands & 4 \\
\midrule
\multicolumn{4}{l}{\textbf{Evolution configuration}} \\
\midrule
Patch types & [diff, full, cross] & Patch type probs & [0.6, 0.3, 0.1] \\
Generations & 75 & Max parallel jobs & 1 \\
Max patch resamples & 3 & Max patch attempts & 3 \\
Meta recommendation interval & 10 & Max meta recommendations & 5 \\
Embedding model & text-embedding-3-small & Max novelty attempts & 3 \\
Code embed sim threshold & 0.95 & Problem implementation & Python \\
LLM dynamic selection & null & Exploration coefficient & 0.0 \\
\midrule
\multicolumn{4}{l}{\textbf{LLM models}} \\
\midrule
gemini-2.5-pro & $\checkmark$ & gemini-2.5-flash & $\times$ \\
claude-sonnet-4 & $\checkmark$ & o4-mini & $\checkmark$ \\
gpt-5 & $\times$ & gpt-5-nano & $\times$ \\
gpt-4.1 & $\times$ & gpt-4.1-mini & $\times$ \\
\midrule
\multicolumn{4}{l}{\textbf{LLM settings}} \\
\midrule
Temperatures & [0.0, 0.5, 1.0] & Max tokens & 16{,}384 \\
Meta models & [gpt-4.1] & Meta temperatures & [0.0] \\
Novelty models & [gpt-4.1] & Novelty temperatures & [0.0] \\
\bottomrule
\end{tabular}
  \caption{\ouralgo Hyperparameter Configuration for the Math Reasoning Agentic Harness.}
  \label{tab:hyperparams_aime}
\end{table}

\clearpage
\subsection{ALE-Bench Problems}

\paragraph{Detailed Task Description.} The ALE-Bench benchmark \citep{imajuku2025ale} is a collection of heuristic programming problems previously used in competitive programming contests (AtCoder). We evaluate \ouralgo on the LITE subset of problems, which consists of 10 problems. We follow the evaluation protocol of ALE-Agent \citep{imajuku2025ale} and use the score calculated on the public test cases as the fitness function. Afterwards, we submit the best solution to the private test set and report the score. Additionally, in \Cref{fig:results_alebench}, we provide scores for evaluating the top-5 publicly scored solutions and taking their maximum score on the private test set. While this does not resemble the traditional competitive programming setting, it allows us to assess the generalization ability of the discovered solutions. The average solution score improves by a neglible amount from 1923.5 to 1927.0. Hence, we do not observe significant evidence for overfitting to the public test cases.

\textbf{\ouralgo's Hyperparameter Configuration.}

\begin{table}[h]
\centering
\small
\begin{tabular}{lclc}
\toprule
\textbf{Parameter} & \textbf{Value} & \textbf{Parameter} & \textbf{Value} \\
\midrule
\multicolumn{4}{l}{\textbf{Database configuration}} \\
\midrule
Archive size & 50 & Elite selection ratio & 0.3 \\
Archive inspirations & 2 & Top-$k$ inspirations & 2 \\
Migration interval & 10 & Migration rate & 0.1 \\
Island elitism & true & Parent selection strategy & weighted \\
Parent selection $\lambda$ & 10.0 & Number of islands & 2 \\
\midrule
\multicolumn{4}{l}{\textbf{Evolution configuration}} \\
\midrule
Patch types & [diff, full, cross] & Patch type probs & [0.6, 0.3, 0.1] \\
Generations & 50 & Max parallel jobs & 1 \\
Max patch resamples & 3 & Max patch attempts & 3 \\
Meta recommendation interval & 5 & Max meta recommendations & 5 \\
Embedding model & None & Max novelty attempts & None \\
Code embed sim threshold & None & Problem implementation & C++ \\
LLM dynamic selection & ucb1 & Exploration coefficient & 1.0 \\
\midrule
\multicolumn{4}{l}{\textbf{LLM models}} \\
\midrule
gemini-2.5-pro & $\checkmark$ & gemini-2.5-flash & $\checkmark$ \\
claude-sonnet-4 & $\checkmark$ & o4-mini & $\checkmark$ \\
gpt-5 & $\checkmark$ & gpt-5-mini & $\checkmark$ \\
gpt-4.1 & $\times$ & gpt-4.1-mini & $\times$ \\
\midrule
\multicolumn{4}{l}{\textbf{LLM settings}} \\
\midrule
Temperatures & [0.0, 0.5, 1.0] & Max tokens & 16{,}384 \\
Meta models & [gpt-5-mini] & Meta temperatures & [0.0] \\
Novelty models & None & Novelty temperatures & None \\
\bottomrule
\end{tabular}
  \caption{\ouralgo Hyperparameter Configuration for the ALE-Bench Problems.}
  \label{tab:hyperparams_ale}
\end{table}

\clearpage
\subsection{Mixture-of-Experts Load Balancing Loss}

\paragraph{Detailed Task Description.}

\begin{table}[h]
\centering
\small
\begin{tabular}{lcc}
\toprule
\textbf{Hyperparameter} &
\textbf{Small MoE (evolution)} &
\textbf{Large MoE (evaluation)} \\
\midrule
\multicolumn{3}{l}{\textbf{Model architecture}} \\
\midrule
Model parameters & 556M & 2.7B \\
Model parameters & 82M & 404M \\
Number of experts ($N_E$) / active per token ($K$) & 64 / 8 & 64 / 8 \\
Hidden size & 512 & 1024 \\
Hidden size in each MoE expert & 384 & 768 \\
Number of hidden layers & 12 & 16 \\
Number of attention heads & 8 & 16 \\
Number of key--value heads & 8 & 8 \\
Head dimension & 128 & 128 \\
Attention bias & false & false \\
Attention dropout & 0.0 & 0.0 \\
Initializer range & 0.02 & 0.02 \\
RoPE $\theta$ & 1{,}000{,}000 & 1{,}000{,}000 \\
Tied word embeddings & true & true \\
Output router logits & true & true \\
Decoder sparse step & 1 & 1 \\
Router auxiliary loss coefficient ($\lambda$) & 0.01 & 0.001, 0.01, 0.1 \\
Computation dtype & bfloat16 & bfloat16 \\
\midrule
\multicolumn{3}{l}{\textbf{Training setup}} \\
\midrule
Optimizer & AdamW & AdamW \\
Learning rate & $1.0\times10^{-3}$ & $3.0\times10^{-4}$ \\
Weight decay & 0.1 & 0.1 \\
Adam parameters $(\beta_1,\beta_2,\epsilon)$ & (0.9, 0.95, $1\!\times\!10^{-8}$) & (0.9, 0.95, $1\!\times\!10^{-8}$) \\
Learning rate scheduler & Cosine decay & Cosine decay \\
Warmup steps & 70 & 490 \\
\midrule
Maximum sequence length & 1024 & 1024 \\
Global train batch size (sequences) & 1024 & 2048 \\
Tokens per training step & 1{,}048{,}576 & 2{,}097{,}152 \\
Maximum steps & 2000 & 14{,}000 \\
Total tokens & 2.10B & 29.36B \\
Dataset & fineweb & fineweb \\
\bottomrule
\end{tabular}
\caption{MoE architectures and training setup.}
\label{tab:moe_hparams}
\end{table}

The Mixture-of-Expert (MoE) architecture \citep{moe_sem_1, moe_sem_2, moe_mod_1_2exp, moe_mod_2_switch_1exp} has been a critical advancement, enabling scaling breakthroughs in large language model training. MoEs are currently ubiquitous amongst modern open and closed-source flagship models \citep{moe_gemini15, guo2025deepseek, llama4, moe_qwen3, moe_gemini25}. The core principle behind the MoE design is to replace traditional large feed-forward residual blocks with ensembles of smaller modules (the ``experts''), which can be efficiently sharded during training and only partially activated during inference~\citep{moe_mod_2_switch_1exp}. Each expert is itself a small feed-forward network $E_{\ell,i}$ located within a larger ensemble of size $N_E$ at layer $\ell$. The router, a layer-specific linear classifier $h_\ell$, selects the top-$K$ most relevant experts for each token, computing only their outputs:
\begin{equation}
\label{eq:moe_layer_extended}
y_\ell(x) = \sum^{N_E}_{i=1} g_{\ell,i}(x)\, E_{\ell,i}(x), \quad
g_{\ell,i}(x) =
\begin{cases}
\frac{e^{h_{\ell,i}(x)}}{\sum_{j \in \mathcal{T}_K(x)} e^{h_{\ell,j}(x)}}, & \text{if } i \in \mathcal{T}_K(x) \\
0, & \text{otherwise}
\end{cases}
\end{equation}
where $\mathcal{T}_K(x)$ denotes the set of indices corresponding to the top-$K$ router logits $h_{\ell,i}(x)$. This sparsely activated design allows different experts to specialize in distinct problem domains, enabling greater efficiency, scalability, and adaptability in handling diverse prompts.

However, due to the non-differentiability of the top-$K$ expert selection operation, it is critical to provide the router with an auxiliary load balancing loss (LBL). The LBL prevents collapse toward uneven token distributions and under-specialized experts. Devising an effective load balancing loss that simultaneously encourages efficiency and expert specialization, without hindering expressivity, remains an open design challenge that has driven much of the recent progress in MoEs~\citep{moe_sem_2, moe_mod_2_switch_1exp, moe_lbl_evo1, moe_lbl_evo2_zloss, moe_lbl_evo3, moe_deepseek_moe, moe_deepseek_demon, moe_similar_olmoe}. Minor design variations have been shown to significantly affect both efficiency and specialization ability~\citep{moe_deepseek_moe, moe_mixtral_lack_spec, moe_chameleon_lack_spec, moe_deepseek_demon_conc, moe_deepseek_demon}. 

One of the most widely adopted designs is the ``global-batch'' LBL introduced by \citet{moe_sem_2}, which underpins several state-of-the-art open models such as Qwen 3~\citep{moe_qwen3}. For a layer $\ell$ with $N_E$ experts, it is defined as:
\begin{equation}
\label{eq:moe_modern_lbl_extended}
L_{LB} = N_E \cdot \frac{1}{L}\sum_{\ell=1}^L \sum_{i=1}^{N_E} f_{\ell,i} \cdot P_{\ell,i},
\end{equation}
where
\[
f_{\ell,i} = \frac{\text{Tokens routed to expert } i}{\text{Total tokens in layer } \ell}, \qquad
P_{\ell,i} = \frac{\sum_{x} h_{\ell,i}(x)}{\sum_{x,j} h_{\ell,j}(x)}.
\]
This formulation encourages token usage across experts to align with the router’s average soft assignment probabilities.

We evaluate \ouralgo by pretraining a MoE model with 556M parameters, $N_E=64$ experts of which only $K=8$ are active for each token, corresponding to 82M sparsely activated parameters per forward pass (excluding embeddings). Training is performed on 2B tokens from fineweb~\citep{fineweb}. For each program, we define a fitness function consisting of the cross-entropy (CE) loss together with an LBL term weighted by $\lambda=0.01$. To additionally measure load imbalance, we track the L1 deviation from a uniform distribution of token allocations:
\begin{equation}
\label{eq:moe_load_imbalance_extended}
L_{\mathrm{imb}}
= \frac{1}{2}\sum_{i=1}^{N_E} \bigl|f_{\ell,i}- \tfrac{1}{N_E}\bigr| \, ,
\end{equation}
with lower values indicating more even load distribution. This grounding provides \ouralgo a two-fold search objective: minimize CE while improving load balance. To avoid local noise affecting the cross-entropy calculations, we average it over the last 10M tokens. The final fitness score used during evolution is then the negated sum of the two:
\begin{equation}
\label{eq:moe_ce_imb_tradeoff}
r= -(L_\mathrm{CE}
+ L_{\mathrm{imb}}).
\end{equation}
Given the expense of pretraining, we run \ouralgo for only 30 iterations, focusing on \texttt{gpt-4.1}, \texttt{gemini-2.5-pro}, and \texttt{claude-sonnet-4}. To evaluate generality, we scale to a larger 2.7B-parameter MoE of which 404M active (excluding embeddings), trained on slightly under 30B fineweb tokens, and compare across three LBL coefficients $\lambda \in \{0.001, 0.01, 0.1\}$. We used \texttt{AdamW}~\citep{adamw} as the optimizer with cosine decay, and linear warmup. As common practice in modern training regimes, we used rotary positional embeddings~\citep{rope}, SwiGLU MLPs~\citep{swiglu}, and half-precision bfloat16 to efficiently keep our model's weights on device. For the small model used during \ouralgo's evolution, we use a batch size of slightly over $1M$ tokens, for 2K steps. For the larger MoE used double the batch size and seven times the total number of steps. After training, we benchmark against the global-batch LBL baseline in terms of perplexity (Figure~\ref{fig:results_moe}, left) and downstream performance across seven standard evaluations: CommonSense QA \citep{moe_bench_1_commonsenseqa}, HellaSwag \citep{moe_bench_2_hellaswag}, OpenBook QA \citep{moe_bench_3_openbook_qa}, PIQA \citep{moe_bench_4_piqa}, SIQA \citep{moe_bench_5_siqa}, 
WinoGrande \citep{moe_bench_6_winogrande}, and ARC \citep{moe_bench_7_arc}, truncating the number of questions to 1000 for large benchmarks as done by~\citep{fineweb}.

\begin{wrapfigure}{r}{0.33\textwidth}
\vspace{-7mm}
\begin{center}
    \includegraphics[width=0.33\textwidth]{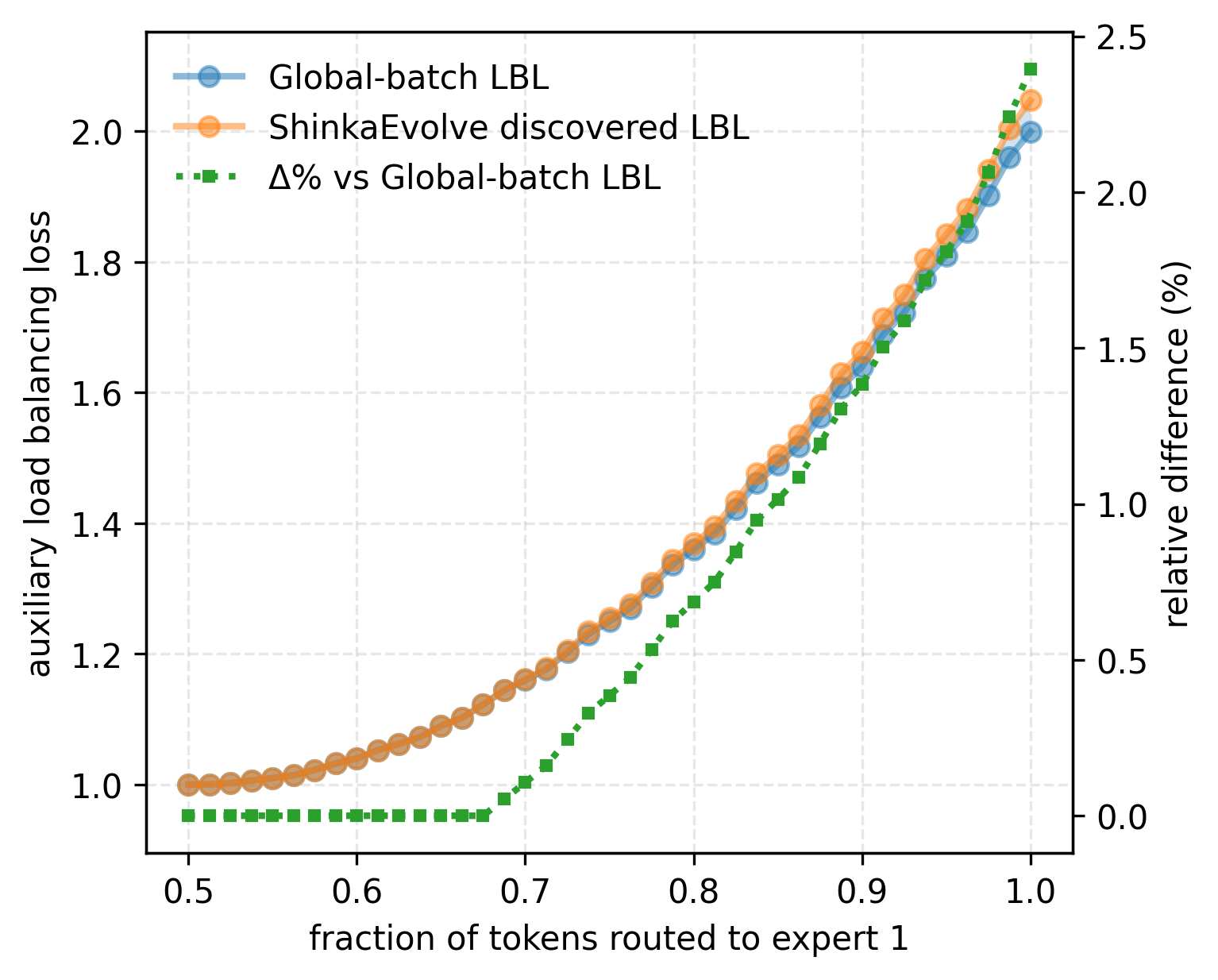}
  \end{center}
  \caption{LBL loss comparison.}
  \label{fig:lbl_loss_value comparison}
  \vspace{-4mm}
\end{wrapfigure}

As described in Section~\ref{sec:results} and detailed in Appendix~\ref{appsec:discovered_solutions}, \ouralgo discovers a new twist on the global-batch LBL from Equation~\ref{eq:moe_modern_lbl_extended}, which was used for seeding evolutionary search. \ouralgo discovers an augmentation of this loss with an additional regularization term to target under-specialized experts. As defined in Equation~\ref{eq:moe_modern_lbl_extended}, let $f_{\ell, i}$ and $P_{\ell, i}$ denote the selection frequency and average router probabilities for expert $i$ in layer $\ell$. Furthermore, define $s(P_\ell)=0.5 + \Bigl(1-\frac{H(P_\ell)}{\log N_E}\Bigr)$ as a normalized complement of the routing entropy, and $\tau=0.064/N_E$ as a minimum usage threshold. The final discovered LBL is:
\begin{equation}
\label{eq:discovered_lbl_extended}
\begin{aligned}
L_{\mathrm{LBL}}
&=
\underbrace{%
\color{lblA} N_E \cdot \frac{1}{L}
\sum_{\ell=1}^{L}\sum_{i=1}^{N_E} f_{\ell,i}\,P_{\ell,i}
}_{\text{Global-batch LBL}}+
\underbrace{%
\color{lblB} \frac{0.1}{L}\sum_{\ell=1}^{L}
s(P_\ell)\,\sum_{i=1}^{N_E}\max\!\bigl(0,\tau-f_{\ell,i}\bigr)
}_{\text{\ouralgo new regularization}} \, .
\end{aligned}
\end{equation}

\begin{figure}[h!]
    \centering
    \includegraphics[width=0.975\textwidth]{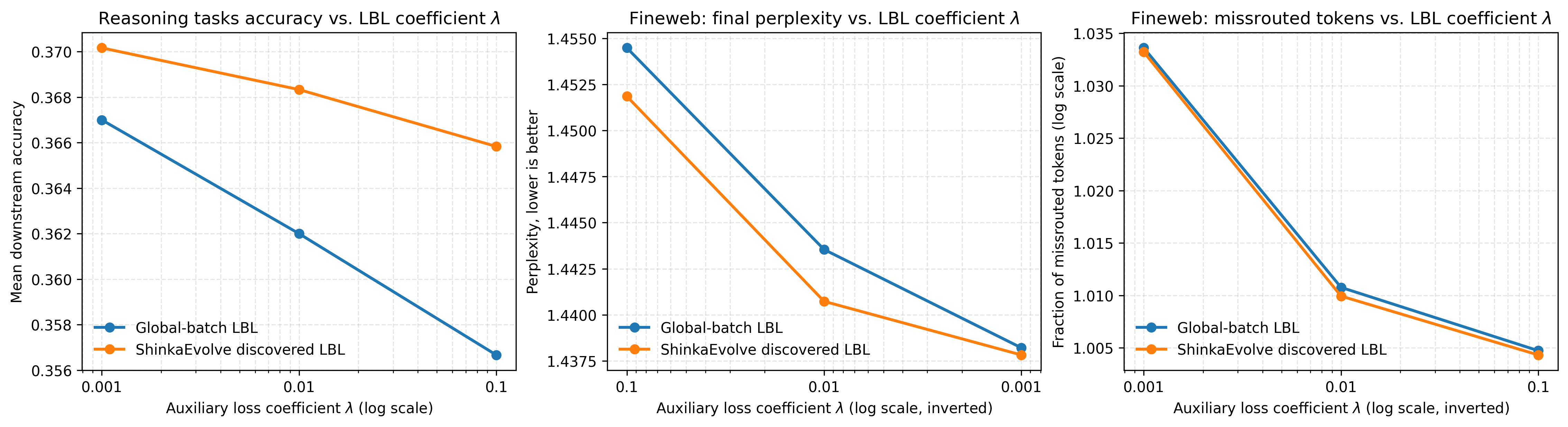}
    \caption{Mixture-of-Experts LBL design additional results.}
    \label{fig:results_moe_app}
\end{figure}
In addition to the results from Section~\ref{sec:results}, in Figure \ref{fig:results_moe_app}, we provide additional results comparing the global-batch LBL and \ouralgo's discovered LBL. In particular, we report the average task performance, final perplexity, and the fraction of missrouted tokens, as a function of the LBL coefficient $\lambda$ used for training the MoEs. Consistent with our previous analysis, \ouralgo's LBL appears to improve from the original LBL across both axes. However, we also note that the architecture used for evolving and testing the employed LBL was quite similar, and the training budget was still limited. However, the consistent generalization results across training budgets and coefficients $\lambda$ provide an optimistic outlook for future extensions to much longer training regimes, where even small efficiency gains could scale to significant cost savings.

\clearpage
\textbf{\ouralgo's Hyperparameter Configuration.}
\begin{table}[h]
\centering
\small
\begin{tabular}{lclc}
\toprule
\textbf{Parameter} & \textbf{Value} & \textbf{Parameter} & \textbf{Value} \\
\midrule
\multicolumn{4}{l}{\textbf{Database configuration}} \\
\midrule
Archive size & 20 & Elite selection ratio & 0.3 \\
Archive inspirations & 4 & Top-$k$ inspirations & 2 \\
Migration interval & 10 & Migration rate & 0.1 \\
Island elitism & true & Parent selection strategy & weighted \\
Parent selection $\lambda$ & 10.0 & Number of islands & 2 \\
\midrule
\multicolumn{4}{l}{\textbf{Evolution configuration}} \\
\midrule
Patch types & [diff, full] & Patch type probs & [0.5, 0.5] \\
Generations & 20 & Max parallel jobs & 1 \\
Max patch resamples & 10 & Max patch attempts & 10 \\
Meta recommendation interval & 10 & Max meta recommendations & 5 \\
Embedding model & text-embedding-3-small & Max novelty attempts & 3 \\
Code embed sim threshold & 0.95 & Problem implementation & Python \\
LLM dynamic selection & ucb1 & Exploration coefficient & 1.0 \\
\midrule
\multicolumn{4}{l}{\textbf{LLM models}} \\
\midrule
gemini-2.5-pro & $\checkmark$ & gemini-2.5-flash & $\times$ \\
claude-sonnet-4 & $\checkmark$ & o4-mini & $\times$ \\
gpt-5 & $\times$ & gpt-5-nano & $\times$ \\
gpt-4.1 & $\checkmark$ & gpt-4.1-mini & $\times$ \\
\midrule
\multicolumn{4}{l}{\textbf{LLM settings}} \\
\midrule
Temperatures & [0.0, 0.5, 1.0] & Max tokens & 16{,}384 \\
Meta models & [gpt-4.1] & Meta temperatures & [0.0] \\
Novelty models & [gpt-4.1] & Novelty temperatures & [0.0] \\
\bottomrule
\end{tabular}
  \caption{\ouralgo Hyperparameter Configuration for the MoE LBL Discovery.}
  \label{tab:hyperparams_moe}
\end{table}

\clearpage

\section{\ouralgo Discovered Solutions}
\label{appsec:discovered_solutions}
\subsection{Circle Packing Problem}
\label{appsec:circle_packing}

\begin{lstlisting}[
  language=Python,
  caption={\ouralgo Discovered Circle Packing Solution.},
  label={lst:circle_packing},
  basicstyle=\ttfamily\tiny  % Change font size here
]
# EVOLVE-BLOCK-START
import numpy as np
from scipy.optimize import minimize, Bounds


def construct_packing():
    """
    Constructs an arrangement of 26 circles by combining a meta-heuristic
    search with a powerful SLSQP optimizer for refinement.
    """
    n = 26

    # --- Helper functions for the optimizer ---
    def objective_func(x):
        """The function to be minimized: the negative sum of radii."""
        return -np.sum(x[:n])

    def constraints_func(x):
        """
        Computes constraint violations. For SLSQP, each value must be >= 0.
        """
        radii = x[:n]
        centers = x[n:].reshape((n, 2))

        containment = np.concatenate(
            [
                centers[:, 0] - radii,
                centers[:, 1] - radii,
                1 - centers[:, 0] - radii,
                1 - centers[:, 1] - radii,
            ]
        )

        overlap = []
        for i in range(n):
            for j in range(i + 1, n):
                dist = np.linalg.norm(centers[i] - centers[j])
                overlap.append(dist - (radii[i] + radii[j]))

        return np.concatenate([containment, np.array(overlap)])

    def _compute_initial_radii(centers):
        """
        Computes a valid set of initial radii for a given set of centers
        to create a feasible starting point (x0) for the optimizer.
        """
        radii = np.min(
            [centers[:, 0], centers[:, 1], 1 - centers[:, 0], 1 - centers[:, 1]], axis=0
        )

        for _ in range(100):
            improved = False
            for i in range(n):
                for j in range(i + 1, n):
                    dist = np.linalg.norm(centers[i] - centers[j])
                    if radii[i] + radii[j] > dist:
                        excess = (radii[i] + radii[j] - dist) * 0.501
                        total_r = radii[i] + radii[j]
                        if total_r > 1e-9:
                            radii[i] -= excess * (radii[i] / total_r)
                            radii[j] -= excess * (radii[j] / total_r)
                            improved = True
            if not improved:
                break
        return np.maximum(radii, 1e-6)

    # --- 1. Generate a single high-quality initial guess ---
    centers_init = np.zeros((n, 2))
    inset = 0.06
    centers_init[0:4] = [
        [inset, inset],
        [1 - inset, inset],
        [inset, 1 - inset],
        [1 - inset, 1 - inset],
    ]
    centers_init[4:8] = [[0.5, inset], [0.5, 1 - inset], [inset, 0.5], [1 - inset, 0.5]]
    centers_init[8] = [0.5, 0.5]

    golden_angle = np.pi * (3 - np.sqrt(5))
    cx, cy = 0.5, 0.5
    inner_r, outer_r = 0.23, 0.48
    inner_idx, outer_idx = np.arange(9, 15), np.arange(15, 26)

    for i, idx in enumerate(inner_idx):
        angle = i * golden_angle
        centers_init[idx] = [cx + inner_r * np.cos(angle), cy + inner_r * np.sin(angle)]
    for i, idx in enumerate(outer_idx):
        angle = i * golden_angle * 1.003
        centers_init[idx] = [cx + outer_r * np.cos(angle), cy + outer_r * np.sin(angle)]

    centers_init += np.random.uniform(
        -0.01, 0.01, size=(n, 2)
    )  # Increased initial jitter
    centers_init = np.clip(centers_init, 0.01, 0.99)

    # --- 2. Define bounds and constraints for the solver ---
    bounds = Bounds([0.0] * n + [0.0] * (2 * n), [0.5] * n + [1.0] * (2 * n))
    constraints = {"type": "ineq", "fun": constraints_func}

    # --- 3. Initial baseline optimization ---
    radii_init = _compute_initial_radii(centers_init)
    x0 = np.concatenate([radii_init, centers_init.flatten()])

    result = minimize(
        objective_func,
        x0,
        method="SLSQP",
        bounds=bounds,
        constraints=constraints,
        options={"maxiter": 600, "ftol": 1e-8, "disp": False},
    )  # Increased initial maxiter

    # Initialize current and best solutions for SA
    best_x = result.x.copy()
    current_x = result.x.copy()
    best_score = -result.fun
    current_score = -result.fun

    # --- 4. Simulated Annealing loop: Perturb and refine with acceptance criterion ---
    sa_iterations = 250  # Significantly increased iterations for SA
    temperature = 0.05  # Initial temperature for SA
    initial_temperature = temperature  # Preserve for potential reheating
    cooling_rate = 0.995  # Slower cooling rate for broader search
    perturb_step = 0.04  # Initial step size for perturbations
    initial_perturb_step = perturb_step  # Preserve for potential reheating
    step_decay = 0.999  # Decay rate for step size
    last_improve = 0  # Iteration of last best improvement
    stagnation_limit = sa_iterations // 4  # Iterations before triggering reheating

    for iter_idx in range(sa_iterations):
        candidate_centers = (
            current_x[n:].reshape((n, 2)).copy()
        )  # Start from current state

        # Select a move type: 70% local, 30% global ring rotation
        if np.random.rand() < 0.7:
            # Local move: perturb a few circles
            num_to_move = np.random.randint(2, 6)
            indices = np.random.choice(n, num_to_move, replace=False)
            candidate_centers[indices] += np.random.normal(
                0, perturb_step, size=(num_to_move, 2)
            )
        else:
            # Global move: rotate one of the rings
            idx_to_rotate = inner_idx if np.random.rand() < 0.5 else outer_idx
            center_point = candidate_centers[8]  # Center of the overall pattern
            angle = np.random.normal(
                0, 0.15
            )  # Angular perturbation (can be fixed or scaled)
            rel_pos = candidate_centers[idx_to_rotate] - center_point
            cos_a, sin_a = np.cos(angle), np.sin(angle)
            rotated = np.column_stack(
                [
                    cos_a * rel_pos[:, 0] - sin_a * rel_pos[:, 1],
                    sin_a * rel_pos[:, 0] + cos_a * rel_pos[:, 1],
                ]
            )
            candidate_centers[idx_to_rotate] = center_point + rotated

        candidate_centers = np.clip(
            candidate_centers, 0.01, 0.99
        )  # Clip to stay within bounds

        # Create a new starting point and run a shorter refinement optimization
        x0_candidate = np.concatenate(
            [_compute_initial_radii(candidate_centers), candidate_centers.flatten()]
        )
        refine_result = minimize(
            objective_func,
            x0_candidate,
            method="SLSQP",
            bounds=bounds,
            constraints=constraints,
            options={"maxiter": 150, "ftol": 1e-6, "disp": False},
        )  # Reduced maxiter, looser ftol

        new_score = -refine_result.fun

        # Simulated Annealing Acceptance Criterion
        # Accept if better, or with probability if worse (based on temperature)
        if new_score > current_score or (
            temperature > 1e-7
            and np.random.rand() < np.exp((new_score - current_score) / temperature)
        ):
            current_score = new_score
            current_x = refine_result.x.copy()  # Update current state
            if new_score > best_score:
                best_score = new_score
                best_x = refine_result.x.copy()  # Update global best
                last_improve = iter_idx  # Reset stagnation counter on improvement
        # If not accepted, current_x remains unchanged for the next iteration (implicit)

        # Cool down temperature and decay perturbation step size
        temperature *= cooling_rate
        perturb_step *= step_decay
        if temperature < 1e-7:
            temperature = 1e-7  # Prevent division by zero
        if perturb_step < 1e-5:
            perturb_step = 1e-5  # Prevent step from becoming too small
        # Reheat if stagnated beyond stagnation_limit
        if iter_idx - last_improve > stagnation_limit:
            temperature = initial_temperature
            perturb_step = initial_perturb_step
            last_improve = iter_idx

    # --- 5. Final Polishing Run on the best found solution ---
    final_result = minimize(
        objective_func,
        best_x,
        method="SLSQP",
        bounds=bounds,
        constraints=constraints,
        options={"maxiter": 1000, "ftol": 1e-9, "disp": False},
    )  # Increased maxiter for final polish

    # Check if the final polishing improved the best_x from SA
    if -final_result.fun > best_score:
        best_x = final_result.x.copy()  # Make sure to copy

    # --- 6. Unpack and return the best result ---
    final_radii = best_x[:n]
    final_centers = best_x[n:].reshape((n, 2))
    return final_centers, final_radii


def compute_max_radii(centers):
    """
    This function is retained for structural compatibility with the evaluation
    framework but is not used by the new `construct_packing` logic.
    It computes maximum radii for a fixed set of centers.
    """
    n = centers.shape[0]
    radii = np.empty(n)
    for i in range(n):
        x, y = centers[i]
        radii[i] = min(x, y, 1 - x, 1 - y)

    for _ in range(60):
        improved = False
        for i in range(n):
            for j in range(i + 1, n):
                dist = np.linalg.norm(centers[i] - centers[j])
                if radii[i] + radii[j] > dist:
                    excess = (radii[i] + radii[j] - dist) * 0.5
                    total = radii[i] + radii[j]
                    if total > 0:
                        reduce_i = excess * (radii[i] / total)
                        reduce_j = excess * (radii[j] / total)
                        radii[i] = max(0.001, radii[i] - reduce_i)
                        radii[j] = max(0.001, radii[j] - reduce_j)
                        improved = True
        if not improved:
            break
    return radii


# EVOLVE-BLOCK-END


# This part remains fixed (not evolved)
def run_packing():
    """Run the circle packing constructor for n=26"""
    np.random.seed(7)
    centers, radii = construct_packing()
    # Calculate the sum of radii
    sum_radii = np.sum(radii)
    return centers, radii, sum_radii


centers, radii, sum_radii = run_packing()
\end{lstlisting}

\clearpage
\subsection{AIME Math Reasoning Agentic Harness}

\label{appsec:aime_agent}

\begin{lstlisting}[
  language=Python,
  caption={\ouralgo Discovered AIME Agent Scaffold Design.},
  label={lst:aime_agent},
  basicstyle=\ttfamily\tiny  % Change font size here
]
"""Agent design evaluation on math tasks."""

import re
from typing import Callable, List, Optional, Tuple, Dict
from collections import Counter, defaultdict
from math_eval import agent_evaluation


# EVOLVE-BLOCK-START
import re
from collections import Counter


class Agent:
    def __init__(
        self,
        query_llm: Callable,
        temperature=0.0,
    ):
        self.query_llm = query_llm
        self.output_format_instructions = "On the final line output only the digits of the answer (0-999). Provide your final answer enclosed in a LaTeX \\boxed{{...}} command."

        # Parameters
        self.generation_temperature = 0.7
        self.review_temperature = 0.1
        self.synthesis_temperature = 0.0

        # Use 3 experts to stay within a 10-call limit (3 gen + 3 review + 1 synth = 7 calls)
        self.num_experts = 3
        self.expert_personas = [
            "You are a meticulous and cautious mathematician. Your guiding principle is 'slow and steady wins the race'. You solve problems by breaking them down into the smallest possible steps based on fundamental principles. You avoid leaps of logic and verify each step before proceeding.",
            "You are a brilliant and intuitive mathematician, known for finding elegant, non-obvious solutions. You look for symmetries, invariants, or a change of perspective that radically simplifies the problem. You trust your insights but explain them clearly.",
            "You are a mathematician with a strong background in computer science. You approach problems by trying to frame them algorithmically. You think in terms of states, transitions, and recurrence relations, and you analyze the behavior of these systems to find the solution.",
        ]

    def _extract_answer(self, text: str) -> Optional[str]:
        """Extracts the final answer from a \\boxed{} environment."""
        if not text:
            return None
        matches = re.findall(r"\\boxed\{(\d{1,3})\}", text)
        if matches:
            return matches[-1]
        return None

    def forward(self, problem: str) -> tuple[str, float]:
        """
        Solves a problem using a multi-persona ensemble with peer review and synthesis.
        """
        total_cost = 0.0

        # === STAGE 1: Generate Diverse Solutions with Expert Personas ===
        solutions = []
        for i in range(self.num_experts):
            persona = self.expert_personas[i % len(self.expert_personas)]
            prompt = f"Solve the following AIME problem by thinking step-by-step. {self.output_format_instructions}\n\nPROBLEM:\n{problem}\n\nSOLUTION:"
            try:
                response, cost = self.query_llm(
                    prompt=prompt,
                    system=persona,
                    temperature=self.generation_temperature,
                )
                solutions.append(response)
                total_cost += cost
            except Exception:
                # If a query fails, we proceed with fewer solutions.
                solutions.append(f"Expert {i + 1} failed to generate a solution.")

        # === STAGE 2: Independent Peer Review & Self-Correction ===
        critiques = []
        reviewer_system_prompt = "You are a skeptical peer reviewer examining a proposed solution to an AIME problem. Your task is to be extremely critical. Do not accept any statement at face value. Re-read the original problem carefully. Check calculations. Scrutinize the logical flow. **Pattern Verification:** If the solution relies on a pattern, you MUST test it on several new examples. If you find an error, clearly explain the flaw and provide a corrected line of reasoning and a final corrected answer. If the solution is completely sound, state that and re-state the final answer."
        for sol in solutions:
            prompt = f"Original Problem:\n{problem}\n\nProposed Solution to Review:\n{sol}\n\nYour Critical Review and Corrected Solution:"
            try:
                review, cost = self.query_llm(
                    prompt=prompt,
                    system=reviewer_system_prompt,
                    temperature=self.review_temperature,
                )
                critiques.append(review)
                total_cost += cost
            except Exception:
                critiques.append("Reviewer failed to provide a critique.")

        # === STAGE 3: Synthesize Final Answer ===
        synthesis_prompt_parts = [
            f"You are the Editor-in-Chief of a prestigious mathematics journal, responsible for publishing the final, canonical solution to this AIME problem. You have received {self.num_experts} independent attempts and their corresponding critical reviews. Your task is to produce the definitive solution.\n\nProblem:\n{problem}"
        ]
        for i, (sol, crit) in enumerate(zip(solutions, critiques)):
            synthesis_prompt_parts.append(
                f"\n--- ATTEMPT {i + 1} ---\nSolution: {sol}\nCritique: {crit}\n---"
            )

        synthesis_prompt_parts.append(
            f"\nSYNTHESIS AND FINAL JUDGEMENT:\n1. First, briefly state the final numerical answer proposed by each of the reviewed attempts.\n2. Based on the critiques, determine which approach is the most reliable, or if all are flawed. Explain your reasoning.\n3. Construct the final, clear, step-by-step, correct solution. Leverage insights from the valid parts of the attempts and correct any identified errors. {self.output_format_instructions}"
        )

        synthesizer_prompt = "\n".join(synthesis_prompt_parts)
        synthesizer_system_prompt = "You are a master mathematician and editor, synthesizing multiple reviewed solutions into one canonical, correct answer."

        final_response = ""
        try:
            final_response, cost = self.query_llm(
                prompt=synthesizer_prompt,
                system=synthesizer_system_prompt,
                temperature=self.synthesis_temperature,
            )
            total_cost += cost
        except Exception:
            pass  # Fallback logic will handle this.

        # === Fallback Logic ===
        if self._extract_answer(final_response) is None:
            # First, trust the reviewed answers
            reviewed_answers = [self._extract_answer(c) for c in critiques]
            valid_reviewed_answers = [
                ans for ans in reviewed_answers if ans is not None
            ]

            if valid_reviewed_answers:
                most_common_answer = Counter(valid_reviewed_answers).most_common(1)[0][
                    0
                ]
                final_response += f"\n\n[Fallback to Majority Vote on Reviewed Solutions]\n\\boxed{{{most_common_answer}}}"
            else:
                # If reviews didn't produce answers, check original solutions
                original_answers = [self._extract_answer(s) for s in solutions]
                valid_original_answers = [
                    ans for ans in original_answers if ans is not None
                ]
                if valid_original_answers:
                    most_common_answer = Counter(valid_original_answers).most_common(1)[
                        0
                    ][0]
                    final_response += f"\n\n[Fallback to Majority Vote on Original Solutions]\n\\boxed{{{most_common_answer}}}"
                else:
                    # Ultimate fallback
                    final_response += "\n\n[Fallback] Could not determine a final answer from any stage.\n\\boxed{000}"

        return final_response, total_cost


# EVOLVE-BLOCK-END


def run_experiment(**kwargs):
    from utils import query_llm, create_call_limited_query_llm
    from functools import partial

    # Create base query_llm function
    base_query_llm = partial(query_llm, model_name=kwargs["model_name"])

    # Wrap it with call limiting (max 10 calls per forward pass)
    limited_query_llm = create_call_limited_query_llm(
        base_query_llm,
        max_calls=kwargs["max_calls"],
    )

    accuracy, cost_total, processed, num_llm_calls, df = agent_evaluation(
        Agent, limited_query_llm, year=kwargs["year"]
    )
    return accuracy, cost_total, processed, num_llm_calls, df
\end{lstlisting}

\clearpage
\subsection{ALE-Bench Problems}

\subsubsection{ALE-Bench LITE task: \texttt{ahc039}}

\lstinputlisting[
  language=C,
  caption={\ouralgo Discovered \texttt{ahc039} Solution.},
  label={lst:ale_ahc039},
  basicstyle=\ttfamily\tiny
]{code/ahc039.cpp}

\clearpage
\subsubsection{ALE-Bench LITE task: \texttt{ahc025}}

\begin{lstlisting}[
  language=C,
  caption={\ouralgo Discovered \texttt{ahc025} Solution.},
  label={lst:ale_ahc025},
  basicstyle=\ttfamily\tiny
]
// EVOLVE-BLOCK-START
#include <iostream>
#include <vector>
#include <string>
#include <numeric>
#include <algorithm>
#include <iomanip>
#include <cmath>
#include <set>
#include <map>
#include <chrono>
#include <random>
#include <unordered_map>

// Timer
std::chrono::steady_clock::time_point program_start_time;
std::chrono::milliseconds time_limit_ms(1850); 

// Global problem parameters
int N_items_global, D_groups_global, Q_total_global;
int queries_made = 0;

std::mt19937 rng_engine;

// Query Manager with optimized caching
class QueryManager {
private:
    int N, Q;
    int& queries_made_ref;
    std::vector<char> cmp1_flat; // flat N*N storage for 1v1 comparisons
    std::unordered_map<uint32_t, char> cmp1v2; // for 1v2 comparisons
    std::mt19937& rng;

    inline uint32_t key1v2(int a, int b, int c) const {
        int mn = std::min(b, c), mx = std::max(b, c);
        return (static_cast<uint32_t>(a) << 16) | (static_cast<uint32_t>(mn) << 8) | static_cast<uint32_t>(mx);
    }

    char perform_query_actual(const std::vector<int>& L_items, const std::vector<int>& R_items) {
        queries_made_ref++;
        std::cout << L_items.size() << " " << R_items.size();
        for (int item_idx : L_items) {
            std::cout << " " << item_idx;
        }
        for (int item_idx : R_items) {
            std::cout << " " << item_idx;
        }
        std::cout << std::endl;

        char result_char;
        std::cin >> result_char;
        return result_char;
    }

public:
    QueryManager(int N_, int Q_, int& qm, std::mt19937& r) : N(N_), Q(Q_), queries_made_ref(qm), rng(r) {
        cmp1_flat.assign(N * N, 0);
        cmp1v2.reserve(N * N / 4 + 10);
    }

    char compare1(int a, int b) {
        if (a == b) return '=';
        int mn = std::min(a, b), mx = std::max(a, b);
        char cached = cmp1_flat[mn * N + mx];
        if (cached != 0) {
            if (a == mn) return cached;
            return (cached == '<' ? '>' : (cached == '>' ? '<' : '='));
        }
        if (queries_made_ref >= Q) return '=';
        
        char res = perform_query_actual({a}, {b});
        if (a == mn) {
            cmp1_flat[mn * N + mx] = res;
        } else {
            if (res == '<') cmp1_flat[mn * N + mx] = '>';
            else if (res == '>') cmp1_flat[mn * N + mx] = '<';
            else cmp1_flat[mn * N + mx] = '=';
        }
        return res;
    }

    char compare1v2(int item_curr, int item_prev, int item_s_aux) {
        if (item_curr == item_prev || item_curr == item_s_aux || item_prev == item_s_aux) {
            if (item_prev == item_s_aux) return compare1(item_curr, item_prev);
            if (item_curr == item_prev) return compare1(item_curr, item_s_aux);
            return compare1(item_curr, item_prev);
        }
        uint32_t key = key1v2(item_curr, item_prev, item_s_aux);
        auto it = cmp1v2.find(key);
        if (it != cmp1v2.end()) return it->second;
        if (queries_made_ref >= Q) return '=';
        char res = perform_query_actual({item_curr}, {item_prev, item_s_aux});
        cmp1v2.emplace(key, res);
        return res;
    }

    void exhaust_queries() {
        if (N >= 2) {
            int a = 0, b = 1;
            while (queries_made_ref < Q) {
                perform_query_actual({a}, {b});
                ++b;
                if (b == a) ++b;
                if (b >= N) { 
                    b = 0; 
                    a = (a + 1) % N; 
                    if (b == a) b = (b + 1) % N; 
                }
            }
        }
    }
};

// Weight estimation module
class WeightEstimator {
private:
    static constexpr long long BASE_WEIGHT = 100000;
    static constexpr int FACTOR_GT = 200;
    static constexpr int FACTOR_LT = 50;
    static constexpr int FACTOR_XJ_FALLBACK = 100;

    QueryManager& qm;
    int N, D, Q;

    double estimate_log2(double val) {
        return (val <= 1.0) ? 0.0 : std::log2(val);
    }

    int calculate_query_cost(int N_val, int k_pivots) {
        if (k_pivots <= 0) return 0;
        if (k_pivots == 1) return std::max(0, N_val - 1);
        double cost = 0;
        cost += k_pivots * estimate_log2(k_pivots);
        for (int j = 2; j < k_pivots; ++j) {
            if (j - 1 > 0) cost += estimate_log2(j - 1);
        }
        cost += (N_val - k_pivots) * estimate_log2(k_pivots);
        return static_cast<int>(std::ceil(cost));
    }

    void merge_sort_pivots(std::vector<int>& pivots, int left, int right) {
        if (left >= right) return;
        int mid = (left + right) / 2;
        merge_sort_pivots(pivots, left, mid);
        merge_sort_pivots(pivots, mid + 1, right);
        
        int n1 = mid - left + 1, n2 = right - mid;
        std::vector<int> L(n1), R(n2);
        for (int i = 0; i < n1; ++i) L[i] = pivots[left + i];
        for (int j = 0; j < n2; ++j) R[j] = pivots[mid + 1 + j];
        
        int i = 0, j = 0, k = left;
        while (i < n1 && j < n2) {
            char cmp = qm.compare1(L[i], R[j]);
            if (cmp == '<' || cmp == '=') pivots[k++] = L[i++];
            else pivots[k++] = R[j++];
        }
        while (i < n1) pivots[k++] = L[i++];
        while (j < n2) pivots[k++] = R[j++];
    }

public:
    WeightEstimator(QueryManager& qm_, int N_, int D_, int Q_) : qm(qm_), N(N_), D(D_), Q(Q_) {}

    std::vector<long long> estimate_weights() {
        std::vector<long long> weights(N, BASE_WEIGHT);
        
        // Determine pivot count
        int k_pivots = (N > 0) ? 1 : 0;
        if (N > 1) {
            for (int k = N; k >= 1; --k) {
                if (calculate_query_cost(N, k) <= Q) {
                    k_pivots = k;
                    break;
                }
            }
        }
        k_pivots = std::min(k_pivots, N);

        if (k_pivots == 0) return weights;

        // Select and sort pivots
        std::vector<int> pivots(k_pivots);
        std::vector<int> indices(N);
        std::iota(indices.begin(), indices.end(), 0);
        std::shuffle(indices.begin(), indices.end(), rng_engine);
        for (int i = 0; i < k_pivots; ++i) pivots[i] = indices[i];

        if (k_pivots >= 2) {
            merge_sort_pivots(pivots, 0, k_pivots - 1);
        }

        // Estimate pivot weights
        if (k_pivots == 1) {
            weights[pivots[0]] = BASE_WEIGHT;
            for (int i = 0; i < N; ++i) {
                if (i == pivots[0]) continue;
                char res = qm.compare1(i, pivots[0]);
                if (res == '=') weights[i] = BASE_WEIGHT;
                else if (res == '<') weights[i] = std::max(1LL, BASE_WEIGHT * FACTOR_LT / 100);
                else weights[i] = std::max(1LL, BASE_WEIGHT * FACTOR_GT / 100);
            }
        } else {
            // Multi-pivot estimation
            weights[pivots[0]] = BASE_WEIGHT;
            
            // Handle p1
            char res_p1 = qm.compare1(pivots[1], pivots[0]);
            if (res_p1 == '=') weights[pivots[1]] = weights[pivots[0]];
            else if (res_p1 == '<') weights[pivots[1]] = std::max(1LL, weights[pivots[0]] * FACTOR_LT / 100);
            else weights[pivots[1]] = std::max(1LL, weights[pivots[0]] * FACTOR_GT / 100);
            
            if (res_p1 == '>' && weights[pivots[1]] == weights[pivots[0]]) {
                weights[pivots[1]] = weights[pivots[0]] + 1;
            }

            // Handle remaining pivots with binary search bracketing
            long long max_bound = BASE_WEIGHT * (N / std::max(1, D) + 10);
            for (int j = 2; j < k_pivots; ++j) {
                int cur = pivots[j], prev = pivots[j-1];
                char res = qm.compare1(cur, prev);
                
                if (res == '=') {
                    weights[cur] = weights[prev];
                } else if (res == '<') {
                    weights[cur] = std::max(1LL, weights[prev] * FACTOR_LT / 100);
                } else {
                    // Binary search to bracket X_j
                    long long X_low = 1, X_high = max_bound;
                    bool low_set = false, high_set = false;
                    
                    int low_idx = 0, high_idx = j - 2;
                    int tries = std::max(1, static_cast<int>(std::ceil(estimate_log2(std::max(1, high_idx - low_idx + 1)))));
                    
                    for (int t = 0; t < tries && low_idx <= high_idx && queries_made < Q; ++t) {
                        int mid_idx = (low_idx + high_idx) / 2;
                        int s = pivots[mid_idx];
                        char res_1v2 = qm.compare1v2(cur, prev, s);
                        
                        if (res_1v2 == '=') {
                            X_low = X_high = weights[s];
                            low_set = high_set = true;
                            break;
                        } else if (res_1v2 == '<') {
                            X_high = weights[s];
                            high_set = true;
                            high_idx = mid_idx - 1;
                        } else {
                            X_low = weights[s];
                            low_set = true;
                            low_idx = mid_idx + 1;
                        }
                    }
                    
                    long long est_X;
                    if (low_set && !high_set) est_X = X_low * FACTOR_GT / 100;
                    else if (!low_set && high_set) est_X = X_high * FACTOR_LT / 100;
                    else if (low_set && high_set) est_X = (X_low + X_high) / 2;
                    else est_X = weights[prev] * FACTOR_XJ_FALLBACK / 100;
                    
                    est_X = std::max(1LL, est_X);
                    weights[cur] = weights[prev] + est_X;
                }
                
                // Ensure monotonicity
                if (weights[cur] < weights[prev]) weights[cur] = weights[prev];
                if (res == '>' && weights[cur] == weights[prev]) weights[cur] = weights[prev] + 1;
            }

            // Estimate non-pivot weights
            std::vector<bool> is_pivot(N, false);
            for (int p : pivots) is_pivot[p] = true;
            
            for (int i = 0; i < N; ++i) {
                if (is_pivot[i]) continue;
                
                int low = 0, high = k_pivots - 1, found = -1;
                while (low <= high && queries_made < Q) {
                    int mid = (low + high) / 2;
                    char res = qm.compare1(i, pivots[mid]);
                    if (res == '=') { found = mid; break; }
                    else if (res == '<') high = mid - 1;
                    else low = mid + 1;
                }
                
                if (found != -1) {
                    weights[i] = weights[pivots[found]];
                    continue;
                }
                
                int pos = low;
                if (pos == 0) {
                    long long w0 = weights[pivots[0]];
                    if (k_pivots >= 2) {
                        long long w1 = weights[pivots[1]];
                        if (w1 > w0 && w0 > 0) weights[i] = std::max(1LL, w0 * w0 / w1);
                        else weights[i] = std::max(1LL, w0 / 2);
                    } else {
                        weights[i] = std::max(1LL, w0 / 2);
                    }
                } else if (pos == k_pivots) {
                    long long wk1 = weights[pivots[k_pivots - 1]];
                    if (k_pivots >= 2) {
                        long long wk2 = weights[pivots[k_pivots - 2]];
                        if (wk1 > wk2 && wk2 > 0) weights[i] = std::max(1LL, wk1 * wk1 / wk2);
                        else weights[i] = std::max(1LL, wk1 * 2);
                    } else {
                        weights[i] = std::max(1LL, wk1 * 2);
                    }
                } else {
                    long long wl = weights[pivots[pos - 1]];
                    long long wr = weights[pivots[pos]];
                    if (wl > 0 && wr > 0) {
                        weights[i] = static_cast<long long>(std::sqrt(static_cast<double>(wl) * wr));
                    } else {
                        weights[i] = (wl + wr) / 2;
                    }
                    weights[i] = std::max(weights[i], wl);
                    weights[i] = std::min(weights[i], wr);
                }
                weights[i] = std::max(1LL, weights[i]);
            }
        }

        // Final validation
        for (int i = 0; i < N; ++i) {
            if (weights[i] <= 0) weights[i] = BASE_WEIGHT;
        }

        return weights;
    }
};

// Assignment optimizer
class AssignmentOptimizer {
private:
    int N, D;
    std::vector<long long>& weights;
    std::mt19937& rng;
    
    double calc_variance(const std::vector<long long>& sums, long long total) {
        if (D <= 0) return 1e18;
        double mean = static_cast<double>(total) / D;
        double sum_sq = 0;
        for (long long s : sums) sum_sq += static_cast<double>(s) * s;
        double var = sum_sq / D - mean * mean;
        return std::max(0.0, var);
    }

public:
    AssignmentOptimizer(int N_, int D_, std::vector<long long>& w, std::mt19937& r) 
        : N(N_), D(D_), weights(w), rng(r) {}

    std::vector<int> optimize() {
        std::vector<int> assignment(N, 0);
        std::vector<long long> group_sums(D, 0);
        std::vector<std::vector<int>> group_items(D);
        std::vector<int> item_pos(N);
        
        // Greedy initialization
        std::vector<std::pair<long long, int>> sorted_items;
        for (int i = 0; i < N; ++i) {
            sorted_items.emplace_back(-weights[i], i);
        }
        std::sort(sorted_items.begin(), sorted_items.end());
        
        long long total_sum = 0;
        for (auto [neg_w, item] : sorted_items) {
            int best_group = 0;
            for (int g = 1; g < D; ++g) {
                if (group_sums[g] < group_sums[best_group]) best_group = g;
            }
            assignment[item] = best_group;
            item_pos[item] = group_items[best_group].size();
            group_items[best_group].push_back(item);
            group_sums[best_group] += weights[item];
            total_sum += weights[item];
        }

        double current_var = calc_variance(group_sums, total_sum);

        // Enhanced local search with best-of-K
        if (D > 1) {
            const int MAX_ITERS = 400;
            const int K_ITEMS = 8;
            
            for (int iter = 0; iter < MAX_ITERS; ++iter) {
                if ((iter & 31) == 0) {
                    auto now = std::chrono::steady_clock::now();
                    if (std::chrono::duration_cast<std::chrono::milliseconds>(now - program_start_time) >= time_limit_ms) break;
                }
                
                int max_g = 0, min_g = 0;
                for (int g = 1; g < D; ++g) {
                    if (group_sums[g] > group_sums[max_g]) max_g = g;
                    if (group_sums[g] < group_sums[min_g]) min_g = g;
                }
                if (max_g == min_g || group_items[max_g].empty()) break;

                // Find best relocate from max_g to min_g among top-K heaviest
                std::vector<std::pair<long long, int>> candidates;
                for (int item : group_items[max_g]) {
                    candidates.emplace_back(weights[item], item);
                }
                if (candidates.empty()) break;
                std::sort(candidates.begin(), candidates.end(), [](const auto& a, const auto& b) { return a.first > b.first; });
                if ((int)candidates.size() > K_ITEMS) candidates.resize(K_ITEMS);

                double best_var = current_var;
                int best_item = -1;
                for (auto [w, item] : candidates) {
                    long long new_max = group_sums[max_g] - w;
                    long long new_min = group_sums[min_g] + w;
                    double new_var = calc_variance({new_max, new_min}, group_sums[max_g] + group_sums[min_g]);
                    if (new_var + 1e-12 < best_var) {
                        best_var = new_var;
                        best_item = item;
                    }
                }

                if (best_item == -1) break;

                // Apply move
                long long w = weights[best_item];
                group_sums[max_g] -= w;
                group_sums[min_g] += w;
                current_var = calc_variance(group_sums, total_sum);

                // Update tracking
                int pos = item_pos[best_item];
                int last = group_items[max_g].back();
                if (best_item != last) {
                    group_items[max_g][pos] = last;
                    item_pos[last] = pos;
                }
                group_items[max_g].pop_back();
                item_pos[best_item] = group_items[min_g].size();
                group_items[min_g].push_back(best_item);
                assignment[best_item] = min_g;
            }
        }

        // Targeted Simulated Annealing
        if (D > 1) {
            double T = std::max(1.0, current_var * 0.25);
            double cool_rate = 0.99985;
            std::uniform_real_distribution<double> unif(0.0, 1.0);
            int iterations = 0, no_imp = 0;
            
            while (true) {
                ++iterations;
                if ((iterations & 255) == 0) {
                    auto now = std::chrono::steady_clock::now();
                    if (std::chrono::duration_cast<std::chrono::milliseconds>(now - program_start_time) >= time_limit_ms) break;
                    T *= cool_rate;
                    if (T < 1e-12) break;
                }

                // Targeted moves: 75% heavy-to-light relocate, 25% swap
                if ((rng() % 4) != 0) {
                    // Targeted relocate
                    int max_g = 0, min_g = 0;
                    for (int g = 1; g < D; ++g) {
                        if (group_sums[g] > group_sums[max_g]) max_g = g;
                        if (group_sums[g] < group_sums[min_g]) min_g = g;
                    }
                    
                    if (group_items[max_g].empty()) { ++no_imp; continue; }
                    
                    // Pick heavy item from max group (best of 3 samples)
                    int item = group_items[max_g][rng() % group_items[max_g].size()];
                    for (int s = 0; s < 2; ++s) {
                        int cand = group_items[max_g][rng() % group_items[max_g].size()];
                        if (weights[cand] > weights[item]) item = cand;
                    }
                    
                    long long w = weights[item];
                    long long new_max = group_sums[max_g] - w;
                    long long new_min = group_sums[min_g] + w;
                    
                    double new_var = current_var;
                    new_var -= (static_cast<double>(group_sums[max_g]) * group_sums[max_g]) / D;
                    new_var -= (static_cast<double>(group_sums[min_g]) * group_sums[min_g]) / D;
                    new_var += (static_cast<double>(new_max) * new_max) / D;
                    new_var += (static_cast<double>(new_min) * new_min) / D;
                    
                    double delta = new_var - current_var;
                    if (delta < 0 || unif(rng) < std::exp(-delta / T)) {
                        // Accept move
                        current_var = new_var;
                        group_sums[max_g] = new_max;
                        group_sums[min_g] = new_min;
                        
                        int pos = item_pos[item];
                        int last = group_items[max_g].back();
                        if (item != last) {
                            group_items[max_g][pos] = last;
                            item_pos[last] = pos;
                        }
                        group_items[max_g].pop_back();
                        item_pos[item] = group_items[min_g].size();
                        group_items[min_g].push_back(item);
                        assignment[item] = min_g;
                        
                        if (delta < -1e-12) no_imp = 0; else ++no_imp;
                    } else ++no_imp;
                } else {
                    // Random swap
                    int g1 = rng() % D, g2 = rng() % D;
                    while (g2 == g1) g2 = rng() % D;
                    if (group_items[g1].empty() || group_items[g2].empty()) { ++no_imp; continue; }
                    
                    int a = group_items[g1][rng() % group_items[g1].size()];
                    int b = group_items[g2][rng() % group_items[g2].size()];
                    long long wa = weights[a], wb = weights[b];
                    
                    long long new_g1 = group_sums[g1] - wa + wb;
                    long long new_g2 = group_sums[g2] - wb + wa;
                    
                    double new_var = current_var;
                    new_var -= (static_cast<double>(group_sums[g1]) * group_sums[g1]) / D;
                    new_var -= (static_cast<double>(group_sums[g2]) * group_sums[g2]) / D;
                    new_var += (static_cast<double>(new_g1) * new_g1) / D;
                    new_var += (static_cast<double>(new_g2) * new_g2) / D;
                    
                    double delta = new_var - current_var;
                    if (delta < 0 || unif(rng) < std::exp(-delta / T)) {
                        current_var = new_var;
                        group_sums[g1] = new_g1;
                        group_sums[g2] = new_g2;
                        
                        // Swap items
                        int pos_a = item_pos[a], pos_b = item_pos[b];
                        int back_a = group_items[g1].back(), back_b = group_items[g2].back();
                        if (a != back_a) { group_items[g1][pos_a] = back_a; item_pos[back_a] = pos_a; }
                        group_items[g1].pop_back();
                        if (b != back_b) { group_items[g2][pos_b] = back_b; item_pos[back_b] = pos_b; }
                        group_items[g2].pop_back();
                        
                        item_pos[b] = group_items[g1].size(); group_items[g1].push_back(b); assignment[b] = g1;
                        item_pos[a] = group_items[g2].size(); group_items[g2].push_back(a); assignment[a] = g2;
                        
                        if (delta < -1e-12) no_imp = 0; else ++no_imp;
                    } else ++no_imp;
                }
                
                if (no_imp > N * 12) break;
            }
        }

        return assignment;
    }
};

int main() {
    std::ios_base::sync_with_stdio(false);
    std::cin.tie(NULL);

    program_start_time = std::chrono::steady_clock::now();
    uint64_t seed = std::chrono::duration_cast<std::chrono::nanoseconds>(
        std::chrono::steady_clock::now().time_since_epoch()).count();
    rng_engine.seed(seed);

    std::cin >> N_items_global >> D_groups_global >> Q_total_global;

    QueryManager qm(N_items_global, Q_total_global, queries_made, rng_engine);
    WeightEstimator estimator(qm, N_items_global, D_groups_global, Q_total_global);
    
    std::vector<long long> weights = estimator.estimate_weights();
    
    qm.exhaust_queries();
    
    AssignmentOptimizer optimizer(N_items_global, D_groups_global, weights, rng_engine);
    std::vector<int> assignment = optimizer.optimize();

    for (int i = 0; i < N_items_global; ++i) {
        std::cout << assignment[i] << (i + 1 == N_items_global ? '\n' : ' ');
    }

    return 0;
}
// EVOLVE-BLOCK-END
\end{lstlisting}

\clearpage
\subsection{Mixture-of-Experts Load Balancing Loss}

\begin{lstlisting}[
  language=Python,
  caption={\ouralgo Discovered Mixture of Experts Load Balancing Loss.},
  label={lst:aime_agent},
  basicstyle=\ttfamily\tiny
]
def load_balancing_loss(
    gate_logits: tuple[torch.Tensor],
    num_experts: int,
    top_k: int = 2,
    attention_mask: Optional[torch.Tensor] = None,
) -> torch.Tensor:
    """
    Load balancing loss for Mixture-of-Experts models.


    parameters
    ----------
    layer_logits:
        list with shape (B, T, total_experts) per layer.
    total_experts:
        number of experts inside the moe feed-forward sub-block.
    top_k_experts:
        number of experts chosen per token (k in top-k gating).
    attention_mask:
        optional mask (B, T) where 0 marks padded tokens.

    returns
    -------
    torch.Tensor:
        scalar loss to be added to the training objective.
    """
    # determine device & flat token count
    device = gate_logits[0].device
    num_layers = len(gate_logits)
    bsz, seqlen = attention_mask.shape
    n_tokens = bsz * seqlen

    # merge layers into (tokens, layers, experts)
    stacked = torch.stack(gate_logits, dim=-2).to(device)
    logits = stacked.view(n_tokens, num_layers, num_experts)

    # obtain routing information
    _, routing_probs, sel_idx = route_logits_to_scores(logits, top_k)
    sel_mask = F.one_hot(sel_idx, num_experts)

    if attention_mask is None:
        # average over all tokens
        avg_sel = sel_mask.float().mean(dim=0)
        avg_prob = routing_probs.mean(dim=0)
    else:
        # expand & apply mask
        m_exp = (
            attention_mask.unsqueeze(-1)
            .unsqueeze(-1)
            .unsqueeze(-1)
            .expand(bsz, seqlen, num_layers, top_k, num_experts)
            .reshape(-1, num_layers, top_k, num_experts)
        )
        avg_sel = sel_mask.float().mul(m_exp).sum(dim=0) / m_exp.sum(dim=0)

        p_mask = (
            attention_mask.unsqueeze(-1)
            .unsqueeze(-1)
            .expand(bsz, seqlen, num_layers, num_experts)
            .reshape(-1, num_layers, num_experts)
        )
        avg_prob = routing_probs.mul(p_mask).sum(dim=0) / p_mask.sum(dim=0)

    # mismatch penalty
    per_layer = avg_sel * avg_prob.unsqueeze(-2)
    main_loss = per_layer.mean(0).sum() * num_experts

    # --- Minimum usage regularizer: softly penalize underused experts ---
    # avg_sel: (layers, top_k, experts)
    # For each expert, sum over top_k to get total selection per expert per layer
    avg_sel_sum = avg_sel.sum(dim=-2)  # (layers, experts)
    # Normalize so that sum over experts = 1 per layer
    avg_sel_norm = avg_sel_sum / (avg_sel_sum.sum(dim=-1, keepdim=True) + 1e-8)

    # Compute entropy of avg_prob per layer (routing distribution)
    entropy = -(avg_prob * torch.log(avg_prob + 1e-8)).sum(dim=-1)  # (layers,)
    max_entropy = torch.log(torch.tensor(num_experts, dtype=avg_prob.dtype, device=avg_prob.device))
    entropy_scale = 1.5 - entropy / (max_entropy + 1e-8)  # ranges from 0.5 (uniform) to 1.5 (concentrated)

    # Penalty: encourage each expert to be used at least min_threshold
    min_threshold = 0.01 * (64.0 / num_experts)
    
    min_usage_penalty = torch.relu(min_threshold - avg_sel_norm).sum(dim=-1)  # (layers,)
    penalty_coeff = 0.1

    # Final loss: main + entropy-scaled min usage penalty
    return main_loss + penalty_coeff * (min_usage_penalty * entropy_scale).mean()
\end{lstlisting}

\end{document}